\documentclass[final,3p,times]{elsarticle}

\usepackage{amssymb}
\usepackage{amsmath}

\usepackage{algorithm}
\usepackage[noend]{algpseudocode}
\usepackage{multirow}
\usepackage{booktabs}
\usepackage{hyperref}
\usepackage{xcolor}
\usepackage{tcolorbox}
\tcbuselibrary{breakable}

\usepackage[multiple]{footmisc}

\journal{}

\makeatletter
\def\ps@pprintTitle{%
     \let\@oddhead\@empty
     \let\@evenhead\@empty
     \def\@oddfoot{\small \textit{Preprint version.} \hfill}%
     \let\@evenfoot\@oddfoot}
\makeatother

\begin{document}

\begin{frontmatter}


\title{RAIDER: Tool-Equipped Large Language Model Agent\\
for \underline{R}obotic \underline{A}ction \underline{I}ssue \underline{D}etection, \underline{E}xplanation and \underline{R}ecovery}


\author[eut,iri]{Silvia Izquierdo-Badiola\corref{cor1}}

\ead{silvia.izquierdo@eurecat.org}
\cortext[cor1]{Corresponding author.}

\author[eut]{Carlos Rizzo}
\author[iri]{Guillem Alenyà}

\affiliation[eut]{organization={Eurecat Centre Tecnològic de Catalunya},
            addressline={Robotics and Automation Unit}, 
            city={Barcelona},
            postcode={08290},
            country={Spain}}
\affiliation[iri]{
            addressline={Institut de Robòtica i Informàtica Industrial CSIC-UPC}, 
            city={Barcelona},
            postcode={08028},
            country={Spain}}

\begin{abstract}

As robots increasingly operate in dynamic human-centric environments, improving their ability to detect, explain, and recover from action-related issues becomes crucial. Traditional model-based and data-driven techniques lack adaptability, while more flexible generative AI methods struggle with grounding extracted information to real-world constraints. We introduce RAIDER, a novel agent that integrates Large Language Models (LLMs) with grounded tools for adaptable and efficient issue detection and explanation. Using a unique ``Ground, Ask\& Answer, Issue" procedure, RAIDER dynamically generates context-aware precondition questions and selects appropriate tools for resolution, achieving targeted information gathering. Our results within a simulated household environment surpass methods relying on predefined models, full scene descriptions, or standalone trained models. Additionally, RAIDER’s explanations enhance recovery success, including cases requiring human interaction. Its modular architecture, featuring self-correction mechanisms, enables straightforward adaptation to diverse scenarios, as demonstrated in a real-world human-assistive task. This showcases RAIDER's potential as a versatile agentic AI solution for robotic issue detection and explanation, while addressing the problem of grounding generative AI for its effective application in embodied agents.
Project website: \href{https://eurecat.github.io/raider-llmagent/}{https://eurecat.github.io/raider-llmagent/}
\end{abstract}






\end{frontmatter}



\section{Introduction}\label{sec:introduction}
``Can you grab the clean kitchen utensil located to the right of the bowl?" To fulfill this request, a human would initially identify the specific kitchen utensil and then assess any obstacles that might hinder the utensil's retrieval. Throughout this process, they would ignore irrelevant details about other objects not related to the task. If any issues impeding the instruction execution were detected, the human would be able to ask for clarification or devise a recovery plan. The question is: how can we empower robots to perform in a similar manner?

Many challenges are faced in developing autonomous robotic systems that can operate and adapt in diverse and dynamic environments. A major challenge consists of the detection, explanation and recovery from issues impeding the execution of a given instruction or action. This work targets two specific issues, namely ambiguity--can the instruction arguments be grounded to specific objects in the scene?—and unfeasibility--is the action feasible without any additional steps? The robot should identify these issues based on the particular instruction and surroundings, interacting with the environment in an efficient and flexible manner. Furthermore, when operating around humans, it should be able to explain the detected issue, for an appropriate recovery plan to be devised.
\begin{figure}[t]
\begin{center}
\includegraphics[width=\columnwidth]{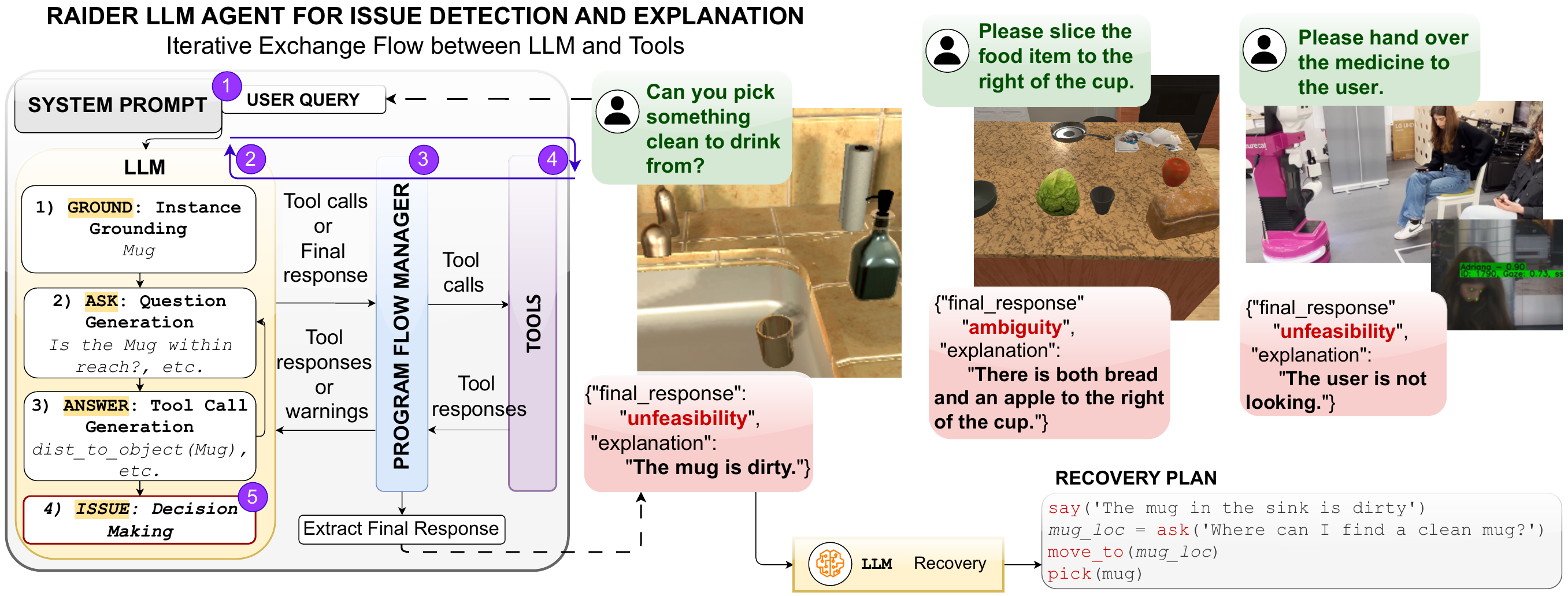}
\end{center}
\caption{RAIDER is composed of an LLM guided by a system prompt and a suite of tools, interconnected through a program flow managing their interactions. The system can process structured or unstructured action queries, referring to objects using various levels of abstraction. Following the \textit{``Ground, Ask\&Answer, Issue''} reasoning procedure, RAIDER dynamically generates and resolves precondition questions to determine \textit{ambiguity} or \textit{unfeasibility} issues. The system's output is leveraged by an LLM to generate a recovery plan involving human interaction.}\label{fig:RAIDER_intro}
\end{figure}

Current methods for action issue detection often rely on predefined precondition checks, requiring extensive action modeling and lacking flexibility to generalize across diverse scenarios without needing to be rebuilt~\cite{ding2023cowp,zhang2023tpvqa,raman2024cape}. 
More flexible techniques using Visual Language Models (VLMs) can address these limitations, but they often provide unreliable answers unless extensively trained on specific datasets, limiting their transferability to different scenarios~\cite{guo2023doremi, skreta2024replan}. Additionally, they rely solely on visual cues, restricting their contextual understanding. Other approaches use Large Language Models (LLMs) to reason about the situation based on complete symbolic scene graphs~\cite{liu2023reflect,goel2024recover}, but processing information that may not be relevant to the specific task might prove inefficient. Furthermore, providing LLMs with irrelevant data can diminish performance, highlighting the need for a more focused strategy that enables an agent to contextually reason about the required information and actively acquire it.
Recent research explores LLM agents that integrate reasoning and external tool use to enhance situational awareness, addressing grounding challenges in embodied AI \cite{shen2023hugginggpt, yang2023LLMGrounder, birr2024autogptp}. We present RAIDER, a system that equips an LLM with a grounded set of tools for efficient and selective information gathering, designed to identify and explain precondition issues in action execution in a zero-shot manner. As illustrated in Figure~\ref{fig:RAIDER_intro}, the LLM-based agent follows a procedure specified in the prompt (step 1) to iteratively reason and interact with external tools (steps 2, 3, 4), and finally determine whether an issue of ambiguity or unfeasibility impedes the execution of the action (step 5). RAIDER's output can then be leveraged by an LLM to find a recovery plan, which may involve interactive actions with the human user as needed.
By focusing on efficiency, adaptability and modularity, our system addresses key limitations in existing approaches.
The system presents the following capabilities, representing our main contributions:
\begin{enumerate}
    \item \textit{Grounding of Abstract Instructions.} The integration of an LLM with grounded tools enables (i) handling of various query abstraction types, overcoming traditional action model limitations, and (ii) more reliable grounding than visual language models on their own.
    \item \textit{Adaptive and Efficient Issue Detection.} Through the procedure \textit{``Ground, Ask\&Answer, Issue"} outlined in the prompt, RAIDER dynamically generates precondition questions based on the context and available tools. Tools are selectively called to gather relevant information without relying on predefined action models, full scene descriptions, or model training, making it adaptable to unseen instructions, while minimizing unnecessary data processing.
    \item \textit{Issue Explanation.} Through contextual information gathering and reasoning, our system produces accurate explanations for detected problems, facilitating recovery and improving its applicability in human-centered contexts.
    \item \textit{Regulated Interaction between LLM and tools.} A program flow manager is designed to inspect the interactions between the LLM and the tools, flagging warnings to target potential LLM hallucinations and tool call errors, enhancing the robustness and transparency of the process.
    \item \textit{Interactive Recovery Plans.} The generated explanations establish a foundation for recovery solutions, involving human interactions when needed.
    \item \textit{Extensible Framework.} RAIDER's modular design allows for extension to different issue detection tasks with minimal reprogramming, by simply updating the prompt's task objective and relevant tools. The rest of components are application-independent, providing a structured and general methodology for reasoning within the context of robotic issue detection.
\end{enumerate}

The aforementioned contributions are supported by the evaluation presented in Section~\ref{sec:evaluation}, demonstrating the system's ability to ground action instructions, detect issues, and explain execution failures across  a number of test cases in a household scenario. Results indicate that our system outperforms methods that utilize full scene descriptions, visual language models, or predefined precondition checks.
Furthermore, the adequacy of RAIDER's explanations for plan recovery is assessed, underscoring their role in successfully generating recovery plans. The section concludes with key findings, including the identified system's strengths and limitations. Section~\ref{sec:extend} explores RAIDER's extensibility across different domains, through toolset modification for handling user preferences and assistive scenarios. A demonstration of RAIDER’s implementation in a real-world assistive plan is presented, highlighting its capability to dynamically detect, explain, and resolve issues related to human factors during plan execution.

\section{Related Work}\label{sec:relatedWork}
\textbf{LLM Agents in Robotics.}
Recent research has focused on developing LLM agents that combine LLMs' reasoning and planning abilities with external tool calling, improving situational awareness and tackling LLMs' grounding challenges. Targeting the reasoning aspect, techniques such as CoT~\cite{wei2022cot} and ToT~\cite{yao2024tot} optimize complex task decomposition, while ReAct~\cite{yao2023react} improves decision-making by intertwining reasoning and action. Self-reflective systems in~\cite{shinn2023reflexion,liu2024raise,liu2024chain,zhao2024expel, madaan2023selfrefine}
use feedback loops and memory to address output hallucinations based on past interactions. However, issues such as long-term memory constraints and dependence on fine-tuning or extensive examples remain prevalent.

When targeting embodied AI, enhancing LLMs with environmental interaction via augmented tools constitutes a promising route. Industry frameworks~\cite{openai_funccall, openai_plugins} and developer tools~\cite{chase2022langchain,nakajima2022babyagi,fernandez2022agentgpt,significantGravitas2020AutoGPT}
exemplify this integration, yet face challenges concerning reliability, abstraction, customization, and troubleshooting. Evaluation benchmarks on LLMs' ability to call functions
reveal inefficiencies, token constraints, and API compliance issues that hinder their seamless integration into embodied systems~\cite{fanjia2024berkeley}.
A number of academic initiatives~\cite{karpas2022MRKL,parisi2022talm,schick2023toolformer,shen2023hugginggpt, patil2024gorilla,Yang2023MMREACTPC}
further explore tool integration with search, calculators, or AI models.

In robotics, LLMs have been used to support task planning from human instructions through frameworks
processing the environment through contextual data~\cite{ahn2022saycan,huang2023innermonologue,liang2022CodeAP,shah2023lmnav}; however, these approaches lack the comprehensive reasoning-with-tools strategy seen in LLM agents. Such agents are starting to be developed in robotics, targeting versatile embodied agents through extensive training~\cite{huang2024leo}, or addressing specific tasks. LLM-Grounder~\cite{yang2023LLMGrounder} enables zero-shot 3D visual grounding with an LLM agent that plans the usage of visual tools, but struggles with cost and latency. The approach presented in~\cite{zhao2023chat} supports tools for robot interaction with objects, uncovering latent object properties for instruction disambiguation. Efforts like AutoGPT+P~\cite{birr2024autogptp} and ``To Help or Not to Help''~\cite{tanneberg2024ToHO} highlight LLM agents in dynamic task planning in social settings, though challenges remain in improving tool selection accuracy and instruction grounding.

While these studies illustrate the potential of LLM agents for complex robotics tasks, issues like hallucinations, incorrect tool selection or prompt specificity hinder real-world effectiveness, calling for further research into framework design and reliable prompt engineering~\cite{valmeekam2022stillcantplan}. Notably, there remain very few examples of LLM agent-based systems applied within real-world robotics. Our work aims to establish an agent framework for zero-shot instruction grounding, issue detection and explanation across diverse robotics scenarios, integrating mechanisms to handle hallucinations and tool inaccuracies while balancing robustness and generalizability.

\textbf{Robotic Issue Detection and Recovery.}
The detection, explanation, and recovery of issues in robotic action execution is a well-researched challenge. This paper primarily targets two specific issues: unfeasibility, where an action cannot be executed due to certain conditions not being met, and ambiguity, pertaining to situations where the action cannot be grounded to specific instances within the environment.

Current strategies for action unfeasibility detection largely rely on predefined models or rules, such as COWP~\cite{ding2023cowp}, TPVQA~\cite{zhang2023tpvqa} and CAPE~\cite{raman2024cape}, which integrate classical planning models with LLMs or VLMs to verify action preconditions and effects and facilitate corrective actions. These techniques often struggle with scalability and dynamic environments, especially when unexpected errors arise. CoPAL~\cite{joublin2024copal} addresses errors with hierarchical replanning but suffers from latency and hard coded precondition checks. Some approaches adopt learned rules, such as the causal-based approach for block stacking~\cite{diehl2023causal}, and DROC~\cite{zha2024droc}, leveraging human feedback to enrich a knowledge base, but they rely on simulation or past human corrections. Alternative strategies, such as Doremi~\cite{guo2023doremi} and RePlan~\cite{skreta2024replan}, combine LLMs with VLMs for action check generation and constraint monitoring. Despite their promise, these approaches can be computationally demanding and may lack reliability due to unfocused diagnostics and dependency on VLM feedback accuracy. Works like REFLECT~\cite{liu2023reflect} and Recover~\cite{goel2024recover} effectively combine symbolic knowledge with LLMs, using scene graphs and ontologies for failure detection and recovery, offering improved explainability. However, their reliance on predefined ontologies and full scene graphs can hinder efficiency and generalizability to novel tasks.

The issue of ambiguity in robotic plan execution has been explored in various studies. KnowNO~\cite{ren2023knowno} utilizes conformal prediction for uncertainty detection in task planning and seeks human clarification when needed. LMCR~\cite{chen2019lmcr} identifies and resolves incomplete instructions via common-sense inference. However, they both lack explanatory feedback. The studies in~\cite{dogan2022followup,hatori2018grndpick,shridhar2018grndreferring} address ambiguities related to referring expressions by generating disambiguation queries to identify the intended objects. While effective, these approaches depend on specialized training and are limited to interpreting spatial relationships, not handling object states or properties.

Our research deals with both ambiguity and unfeasibility issues in a unified framework, by equipping an LLM with grounded tools for dynamic precondition generation and context-aware processing. Unlike prior methods, this approach does not necessitate training or reliance on predefined action models; instead, it engages in a guided analysis of the environment rather than indiscriminately processing the entire scene. Additionally, it considers human involvement in the recovery process when tasks exceed the system's capabilities, a factor often overlooked by existing approaches.

\textbf{Robotic Instruction Grounding.}
For a robot to effectively execute a natural language command, it must anchor the instruction within its environmental context. This entails mapping verbal commands--often rich in semantic relationships, properties, and states--to precise actions or objects within the scene. The challenge of instruction grounding is intertwined with the issue detection problem: no objects matching the command results in unfeasibility, while multiple matches lead to ambiguity. Hence, while not the main focus of our work, we present relevant research focusing on grounding.

Neural network architectures have been developed for object grounding, integrating LSTMs and CNNs to match referring natural language expressions to specific objects in the scene~\cite{magassouba2019targetsourcemodel,shridhar2018grndreferring,hatori2018grndpick}. MOCA's approach in instruction following utilizes dual heatmaps for instance association, marrying perception with policy~\cite{singh2021moca}. Other methods leverage explainability to find defined scene regions, utilizing Grad-CAM and K-means clustering~\cite{fethiye2023explainground}. Visual language models~\cite{liu2024llavanext,radford2021clip} have also been tested as an end-to-end approach for visual grounding. The aforementioned approaches require extensive data collection and specialized fine-tuning to effectively manage object properties, spatial relationships, and states~\cite{yang2023LLMGrounder}, with no individual work covering all.

Many robotic systems now employ LLMs for interpreting instructions; these models, however, lack intrinsic grounding capabilities and depend on external models to interpret sensory data. As presented earlier, a series of works combine LLMs' high-level reasoning to guide the use of external grounded modules. Within task planning contexts, CaP~\cite{liang2022CodeAP} and Instruct2Act~\cite{huang2023instruct2act} leverage code generation to handle ambiguous instructions, through API calls to perception models. Specific to grounding, Zhao\emph{et~al.}~\cite{zhao2023chat} achieve LLM-based interactive perception to uncover latent object properties, albeit within constrained task setups. LLM-Grounder~\cite{yang2023LLMGrounder} implements an LLM agent to decompose the instruction before querying visual models for grounding, but faces high cost and latency issues.

Our approach to grounding aligns with this trajectory, advocating for a streamlined approach that takes advantage of the developed LLM agent capabilities, coupling reasoning with information acquisition tools for zero-shot grounding to the available objects in the scene.

\section{RAIDER Framework}\label{sec:methodology}
\begin{figure}[t]
\begin{center}
\includegraphics[width=1\textwidth]{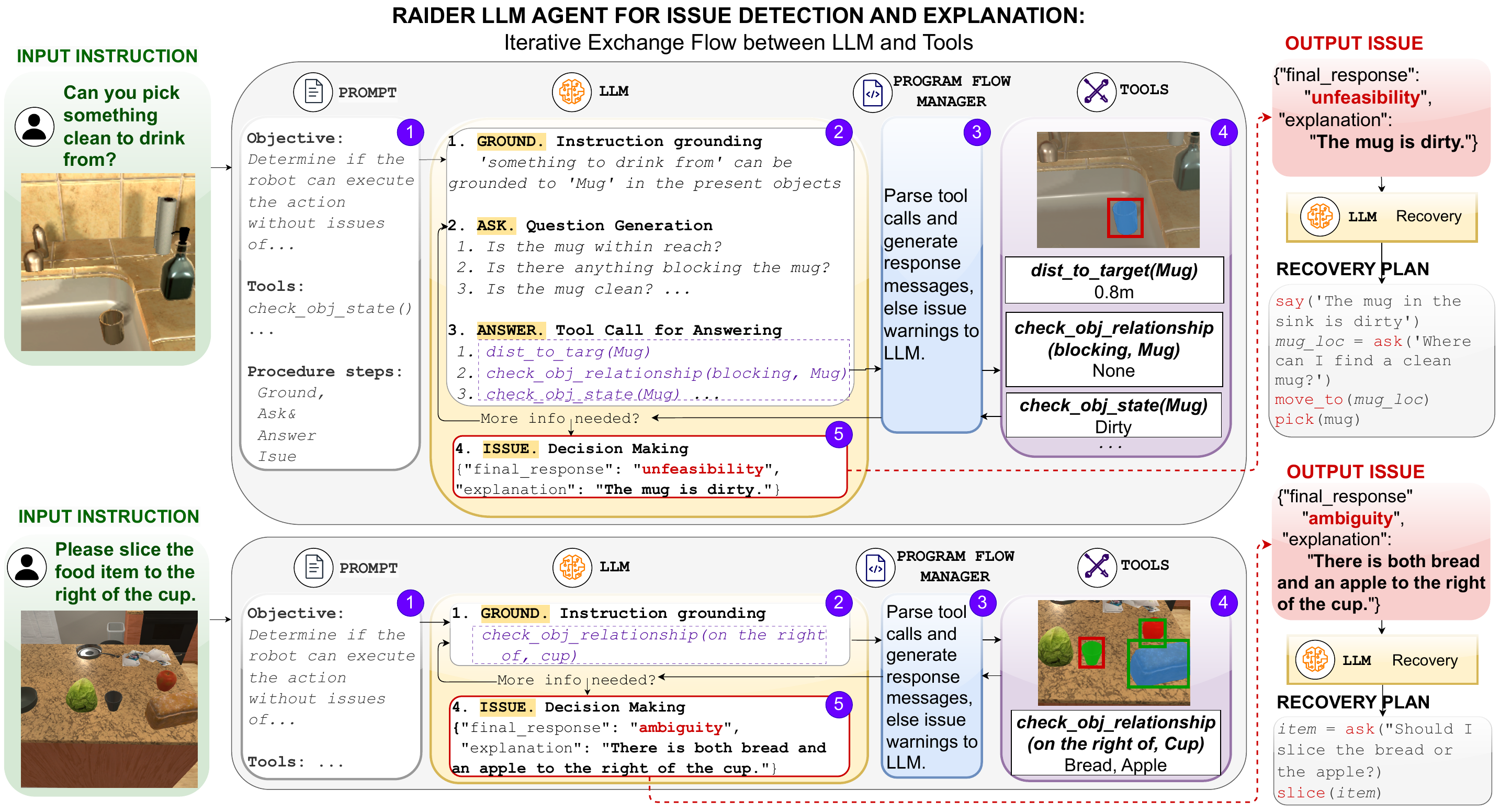}
\end{center}
\caption{RAIDER-LLM Agent. Qualitative examples showing an unfeasibility detection (top) and ambiguity detection (bottom) in action execution. Based on the instruction, available objects and tools, the LLM employs \textit{``Ground, Ask\&Answer, Issue''} reasoning procedure detailed in the prompt, interacting with tools via the program flow manager.}\label{fig:llmagent_ex}
\end{figure}

RAIDER LLM agent is designed to process a query and assess potential ambiguity or unfeasibility issues impeding its execution by the robot. Queries can be expressed in natural language, e.g., (\textit{``Can you slice...?" ``Please open..."}) or as structured actions (e.g., \textit{pick(obj, loc)}). Furthermore, the query might refer to objects using various levels of abstraction, such as specific names (e.g.  \textit{apple}), abstract terms (e.g.  \textit{fruit}), spatial relations (e.g., \textit{to the right of}), or state conditions (e.g., \textit{clean}). Two qualitative examples are illustrated in Figure~\ref{fig:llmagent_ex}--by dynamically interacting with the environment using the available tools, the agent grounds the instruction and delivers a final response comprising the identified issue along with an explanation. An issue of unfeasibility indicates an action that cannot be carried out without further steps due to some conditions not being met, while ambiguity points to multiple objects corresponding to the request.
\subsection{Key Elements}\label{sec:keyelements}

At its core, the agent comprises an LLM, guided by a prompt and a set of tools, interconnected through a program flow that manages their interactions. The LLM reasons about the information needed based on the current context, following the procedure \textit{``Ground, Ask\&Answer, Issue"}, outlined in the system prompt. The prompt also contains the task objective and lists the available tools, which provide reliable and grounded information upon request. The Program Flow Manager regulates the communication between the tools and the LLM, parsing and inspecting messages to flag potential errors, until a final response is generated.

Our system utilizes OpenAI's GPT-4o (\textit{gpt-4o-2024-05-13}). However, the architecture is designed to accommodate any LLM. The following sections provide a detailed description of the system's components.

\subsection{Agent Program Flow Manager - PFM}\label{sec:programflow}
\begin{figure}[!t]
\begin{center}
\includegraphics[width=1\textwidth]{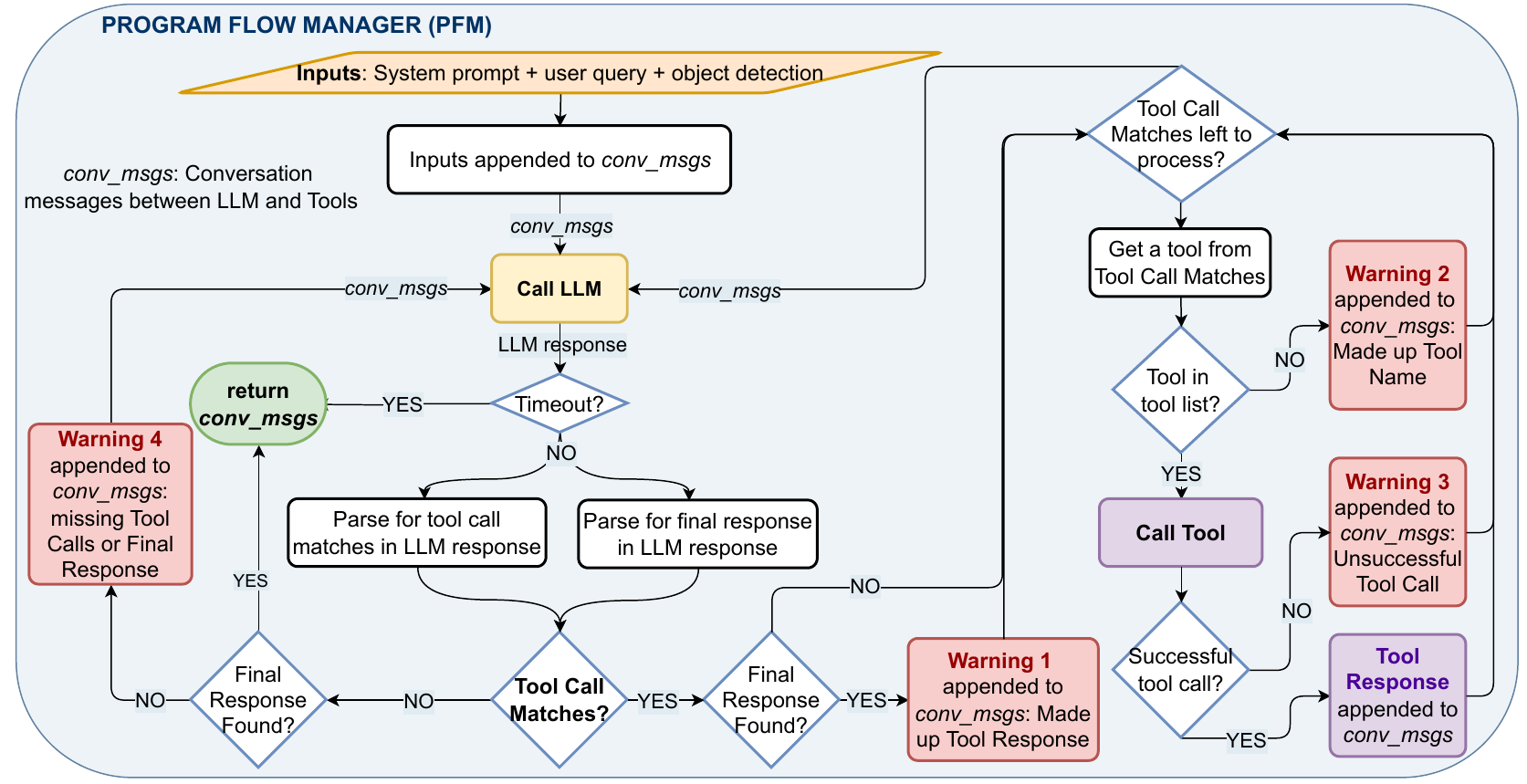}
\end{center}
\caption{The Program Flow Manager (PFM) regulates interactions between the LLM and the tools until a final response is produced with no tool calls left, or a timeout is reached. The interactions are verified at critical steps, with warnings appended to the conversation as required.}\label{fig:pfmdiagram}
\end{figure}

Without the ability to self-evaluate, the agent could easily get stuck in a endless loop, failing to accomplish the given task or produce acceptable outcomes. The agent's program flow manager (PFM) has been developed to regulate the communication between the LLM and its tools, through a number of checks implemented in the iterative program flow outlined in Figure~\ref{fig:pfmdiagram}. The process begins with the agent receiving an input message composed of the system prompt, user query, and the list of objects present in the scene. The input is forwarded to the LLM to generate an initial response, after which the agent enters a loop for continuous assessment of tool calls and LLM responses. This loop will terminate when the LLM provides a final response without any remaining tool calls, or when a timeout occurs. During each iteration, the agent looks for tool call matches or a final response in the LLM's response. If tool call matches are identified, the system will proceed to processing the tool calls sequentially. A warning (Warning 1) is appended to the conversation if a final response is present alongside the pending tool calls, suggesting that the LLM might have made up the answer. For every tool in the matches, if it is listed among the available options, its corresponding call is executed, yielding either a successful response or an error warning (Warning 3), appended to the conversation. If the tool is not in the tool list, a warning (Warning 2) is appended to the conversation messages. This process is repeated until all tool call matches have been processed. On the contrary, if no matches are detected in the LLM's response, the agent returns the conversation messages containing the final response if available, or generates a warning message (Warning 4) indicating that it is still awaiting a final response or additional tool calls. All warning messages and tool responses are systematically appended to the conversation log to keep a memory of prior tool calls and observations, and passed to the LLM in the subsequent iteration for continued processing. If the process exceeds a specified time limit without a final response (20 seconds in our experiments), the agent will terminate and return the collected messages contained in $conv\_msgs$.

As described, the flow incorporates a proactive error-checking mechanism to address potential LLM hallucinations and incorrect requests, thereby enhancing transparency throughout the interaction. When a possible error is identified, one of the following warning messages is generated and appended to the conversation, so that it can be passed back to the LLM for self-reflection and response refinement (see Figure~\ref{fig:warnings} in the appendix for illustrative examples):
\begin{itemize}
    \setlength{\itemsep}{1pt}
    \setlength{\parskip}{1pt}
    \setlength{\parsep}{0pt}
    \item Warning 1: \textit{Made up Tool Response} is triggered when a final response coincides with pending tool calls, indicating that the LLM may have fabricated the tool responses or the final response.
    \item Warning 2: \textit{Made up Tool Name} arises when a tool is not found in the list of available tools, implying that the LLM has fabricated the tool.
    \item Warning 3: \textit{Unsuccessful Tool Call} occurs when a tool fails to yield a valid response, suggesting that the tool call may not have been correctly constructed.
    \item Warning 4: \textit{Missing Tool Call or Final Response} is generated when neither a final response nor correct tool calls are present in the LLM's response, indicating a need for these elements.
\end{itemize}
Notably, this program is application-independent, allowing for any tailored selection of tools and language model. It does not rely on any external agent frameworks, providing complete control over the flow of interactions. This increases performance and avoids high levels of abstraction layers that lead to unexpected behaviors and inconsistencies.

\subsection{System Prompt - Ground, Ask\&Answer, Issue}\label{sec:prompt}
\begin{figure}[t]
\centering
\includegraphics[width=0.93\columnwidth]{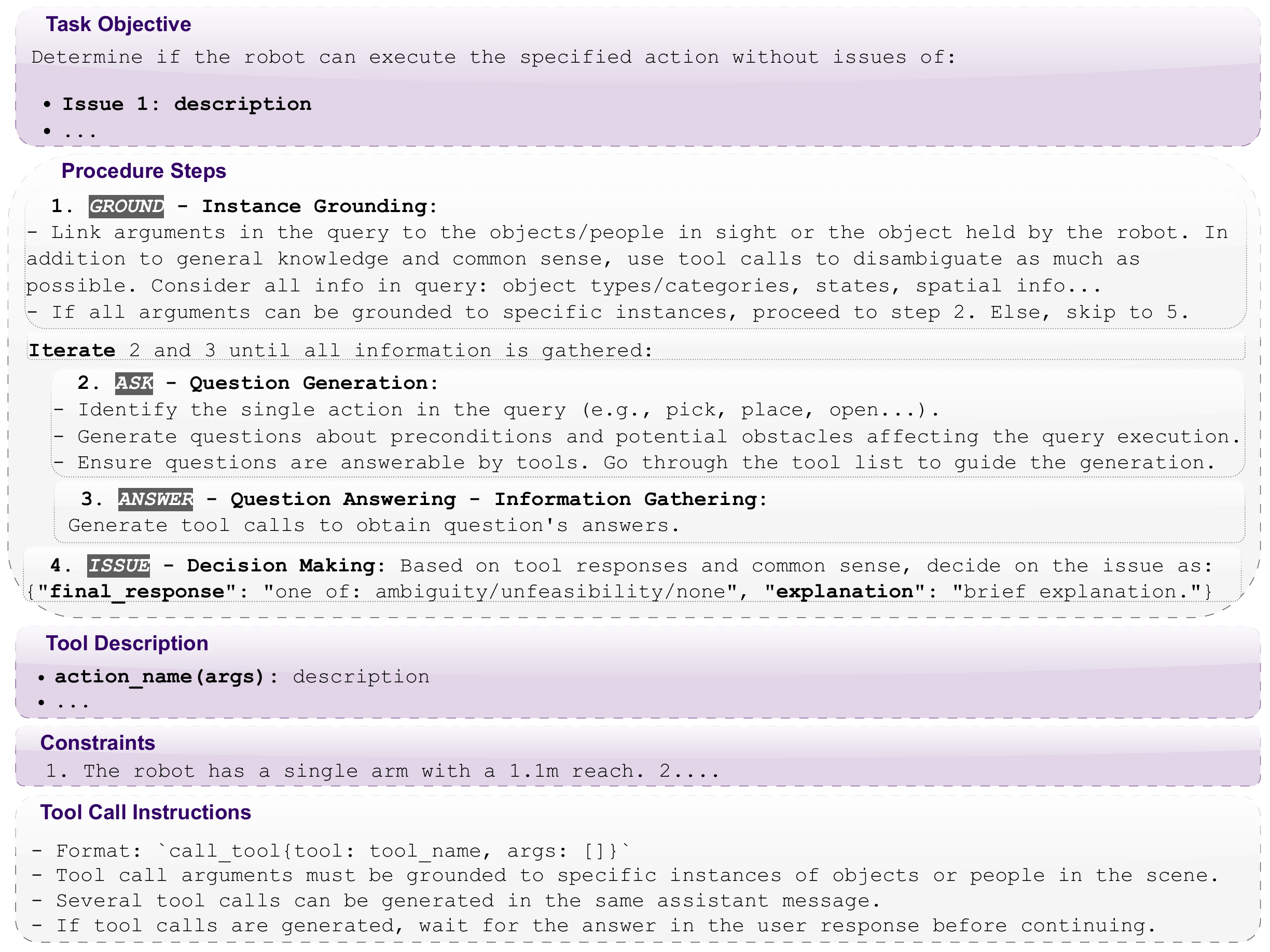}
\caption{Full prompt template for issue detection in robotic action execution. The gray blocks are general to any issue detection application, while the purple blocks should be tailored to each specific scenario.}\label{fig:prompt_templ}
\end{figure}
The system prompt provided to the LLM constitutes a critical component to ensure accurate and high quality responses from the model. It encapsulates the objective of the task, instructions for its execution, constraints that must be adhered to, descriptions of available tools, and guidelines for tool calling. Guided by the prompt, the LLM decides which tools and arguments to invoke throughout the process to solve the task at hand. The prompt designed in this work is aimed at issue detection and explanation in robotic action execution, and is intended to be generalizable to a number of different scenarios and issue types. 
A significant amount of prompt engineering went into the process of designing the final prompt structure, composed of the sections illustrated in Figure~\ref{fig:prompt_templ}. To enable the agent to operate in a new issue detection environment, the task objective, tool description and constraints would need to be tailored to the new application, while the procedure steps and tool call format would remain unchanged.

\begin{figure}[!h]
\centering
\includegraphics[width=0.93\columnwidth]{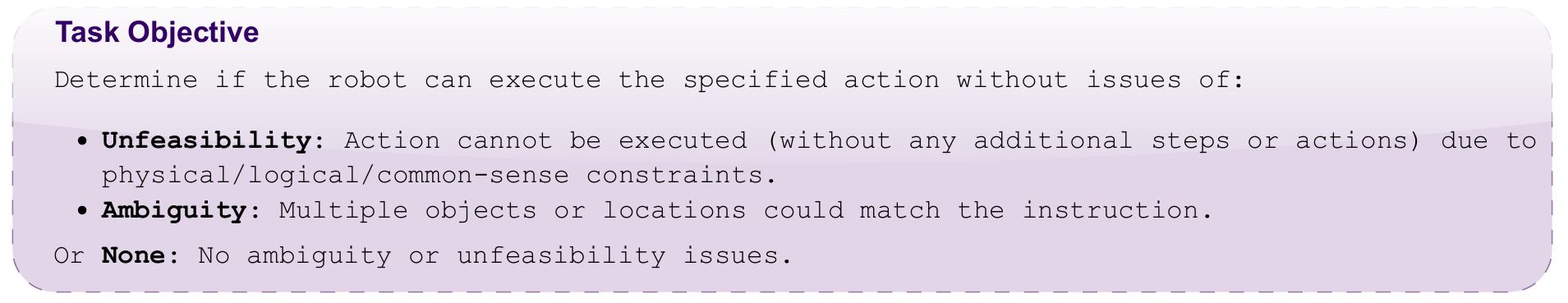}
\caption{Prompt - Task Objective. The task objective consists of a list of potential issues to identify, which can be adapted based on the application needs.}\label{fig:prompttask}
\end{figure}

\textbf{Task Objective.} The task objective is given as a list of potential issues that need to be detected. In this case, we address issues of unfeasibility or ambiguity when executing an action, as described in Figure~\ref{fig:prompttask}. These issues can be modified based on the application requirements, such as detecting if the task entails a risk of harm to nearby individuals, as illustrated in Section~\ref{sec:extend_assist}.

\textbf{Procedure Steps.} A procedure named \textit{``Ground, Ask\&Answer, Issue"} has been designed as a general guideline to direct an LLM through the process of issue detection for a given instruction (see Figure~\ref{fig:prompt_templ}). Action instructions might be provided at different levels of abstraction, ranging from structured actions such as \textit{``pick(apple)"}, to unstructured natural language queries e.g., \textit{``please pick the fruit that is on the left of the bowl"}. Hence, the initial step involves grounding the instruction to an action with arguments linked to specific objects in the scene. Once the query has been grounded, the system is intended to generate questions regarding factors that might impede the successful action execution, i.e. action preconditions. The question generation stage serves as a means for the LLM to consider what information needs to be gathered. Since the questions must be addressed by the tools, the question generation step is influenced by the available tools, which constrain the scope of the agent. In the third step, the LLM generates the corresponding tool calls to answer the generated questions, and waits for tool returns. The agent can then iterate through steps 2 and 3 until all required information has been gathered and a final response can be delivered. This iteration constitutes a version of the ReAct~\cite{yao2023react} \textit{``thought, action, observation"} cycle, adapted to guide and better fit the problem of reasoning and acquiring the required information for robotic issue detection. The added steps for grounding and question generation have shown to be crucial for the system's accuracy, as evidenced by the ablation experiments presented in Section~\ref{sec:evalpromptabl}.

\textbf{Tool Description.} The available tools along with their descriptions are enumerated in this part, with descriptions including their purpose, input arguments and output. The tools defined for the issue detection application in this work are described in the subsequent Section~\ref{sec:tools}

\textbf{Constraints.} This segment should outline the application-specific constraints, encompassing both robot and environment information, along with application-specific guidelines. In our scenario, we specify that the robot is equipped with a single arm that has a reach of 1.1m. Additionally, a line to help question generation is provided: ``Check for environment issues
(wrong object properties, wrong object states, blockages)
and robot issues (wrong states, physical constraints)".

\textbf{Tool Call Format.} Instructions regarding the format in which the tools should be invoked are provided so that tool calls can be parsed by the program flow manager: \textit{`call\_tool\{tool: tool\_name, args: arg\_list\}'}.

\subsection{Tools - Agent's Information Gathering Capabilities}\label{sec:tools}
In the context of LLM agents, tools play a crucial role in the system, as they determine which information can be collected by the agent, establishing both its capabilities and limitations. While the system is not limited by defined actions, it is restricted by the scope of the tools. These should therefore be defined and implemented based on the specific needs and functionalities of the application. A list of these tools and their description is included in the prompt, guiding the agent's behavior for generating questions regarding what to check on.
\begin{figure}[!ht]
\begin{center}
\includegraphics[width=0.9\columnwidth]{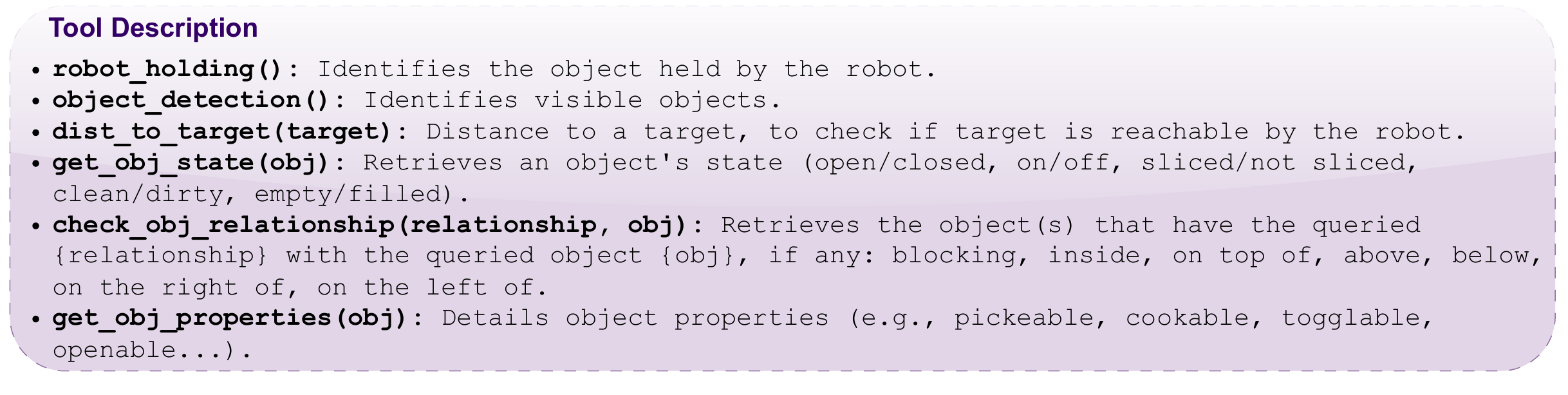}
\end{center}
\caption{Prompt - Tool Description. The tool list is provided in the prompt, including the expected arguments, description and output for each tool.}\label{fig:prompttools}
\end{figure}
For our issue detection application focused on robotic actions in a household environment, we have defined and implemented the tools depicted in Figure~\ref{fig:prompttools}. These tools can be implemented using various perception models and techniques. Further details of our implementation are given in Section~\ref{sec:tool_impl}. Notably, any set of tools can be integrated into the system based on the application needs, with the descriptive prompt modified accordingly, as illustrated in Section~\ref{sec:extend}. The advantage of this system, compared to merely invoking all available tools and supplying all the information, lies in the LLM's ability to determine not only which tools to utilize but also the specific arguments to use when calling them. This significantly reduces the number of tool calls by focusing on relevant information instead of addressing every object in the scene, minimizing unnecessary data processing.


\subsection{RAIDER for Interactive Action Recovery}\label{sec:method_recovery}
\begin{figure}[!h]
\begin{center}
\includegraphics[width=0.9\columnwidth]{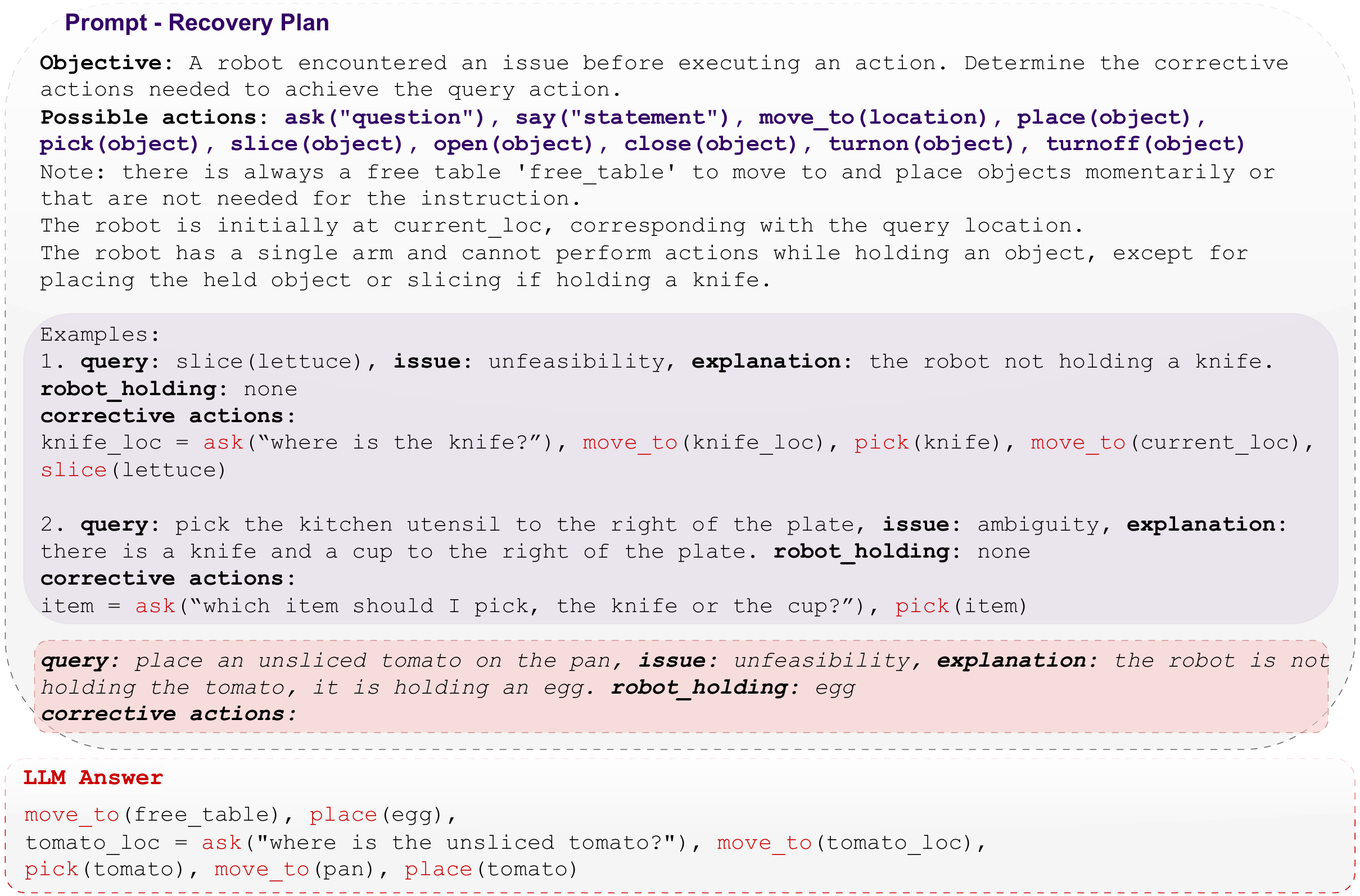}
\end{center}
\caption{Prompt for Interactive Plan Recovery through few-shot prompting, containing the objective, the available actions to the robot, and contextual notes. The query includes the action, the issue and its explanation (RAIDER's output), and the object held by the robot. The resulting plan is formatted as code, so that the output from the interaction actions can be used in posterior actions. The plan might involve the robot solving the task on its own, asking for information or clarification, or stating why an action is unfeasible.}\label{fig:recovllm}
\end{figure}

Once an issue has been identified by the system, RAIDER's output can also be leveraged to facilitate the generation of a recovery plan enabling the execution of the action for which the issue was detected. We implement a recovery LLM that ingests the identified issue and its corresponding explanation, producing a recovery plan through few-shot prompting~\cite{brown2020fewshot}. In addition to a number of examples, the prompt is composed of the objective, the available actions that the robot can take, and a contextual note with relevant information to solve the task (Figure~\ref{fig:recovllm}). We explicitly want to tackle applications entailing HRI, recognizing that the robot might not always be able to autonomously resolve the issue (e.g. in case of unknown object location), and highlighting the necessity for interactive plan recovery. The plan might therefore, in addition to regular actions such as \textit{move} or \textit{place}, include interaction actions: \textit{say} and \textit{ask}. In order to handle the outputs from these interactions for posterior actions, we exploit the LLM's ability to produce code previously demonstrated in literature~\cite{liang2022CodeAP,huang2023instruct2act}. A recovery plan is generated with actions expressed as Pythonic functions, where the interaction action \textit{ask} returns an output to be used in the rest of the plan. With this simple yet efficient approach, the robot is able to either solve the task on its own (e.g. unblocking an object through regular actions \textit{move, pick, place...}), ask for information or clarification when needed (e.g. \textit{ask("Which vegetable should I pick?")}), or state that an action is unfeasible, and why (e.g. \textit{say("I cannot pick the microwave as it is too heavy.")}).

\section{Evaluation}\label{sec:evaluation}
We systematically evaluate RAIDER's ability to ground action instructions and detect and explain potential issues for their execution. To illustrate the benefits of our framework, we conduct comparisons with a number of baseline methods (see Section~\ref{sec:evalmethods}), as well as prompt ablations within the system, described in Section~\ref{sec:evalpromptabl}. An analysis of RAIDER's failure reasons is included in Section~\ref{sec:eval_failures}, and Section~\ref{sec:evalrep} examines the repeatability of the system. Finally, for correctly detected issues, we assess whether a correct recovery plan can be generated from the output issue explanation (Section~\ref{sec:evalrecov}). We conclude with a number of key findings in Section~\ref{sec:key_find}, representing RAIDER's identified strengths and limitations. 
The following metrics are evaluated in our experiments:
\begin{itemize}
    \setlength{\itemsep}{1pt}
    \setlength{\parskip}{1pt}
    \setlength{\parsep}{0pt}
    \item \textit{Grnd} (grounding success): percentage of cases requiring grounding where either the action query is successfully grounded to specific objects or an ambiguity is correctly detected.
    \item \textit{Det} (issue detection success): percentage of cases where the issue is correctly identified.
    \item \textit{Expl} (explanation success): percentage of cases where a correct issue explanation is provided.
    \item \textit{Time} (time taken): time taken to detect and explain the issue, in seconds.
    \item \textit{Recov\_Plan} (recovery plan): percentage of cases where a valid recovery plan is found from the issue explanation provided by the agent.
\end{itemize}
To enable thorough evaluation, this section's assessment is conducted within a simulated environment, as described next in Section~\ref{sec:dataset}. We conduct a number of experiments using real-world images, with some examples provided in the appendix (Figure~\ref{fig:example_real}). Additionally, a demonstration of RAIDER's application in a real-world human assistive scenario is detailed in Section~\ref{sec:extend_assist}.
\subsection{Experimental Setup}

\subsubsection{Dataset}\label{sec:dataset}
The system is extensively evaluated on simulation data from AI2THOR~\cite{kolve_ai2-thor_2017}, using task execution events generated in Liu et al.'s work~\cite{liu2023reflect}. We define 260 test cases, consisting of an action query and an associated AI2THOR event. The event consists of an RGB-D observation as well as ground truth metadata obtained from simulation, used to provide the information requested by the tools as described in Section~\ref{sec:tool_impl}. Events comprise a different number of objects, ranging from 1 to 24. Each test query is categorized by a query type and query abstraction level, distributed in the dataset as shown in Figure~\ref{fig:querytypes}. The queries correspond to a single action primitive, such as pick, open, turn off, etc. Notably, the supported actions are not limited to the ones listed here as they do not need to be predefined or modeled, with the executed checks depending on the semantic meaning of the action query, the context and the available tools. The established categories for query types, query abstraction types, and issue types are as follows:
\begin{figure*}[!h]
\begin{center}
\includegraphics[width=1\textwidth]{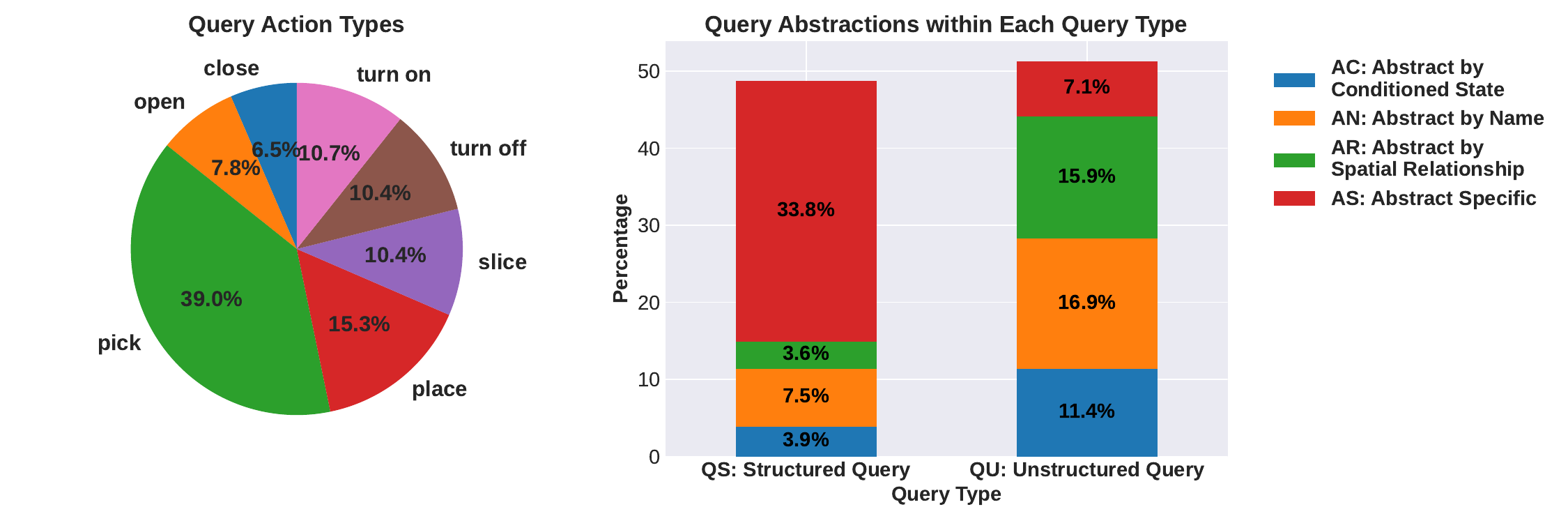}
\end{center}
\caption{Dataset distribution by query actions (left) and query type and abstraction (right).}\label{fig:querytypes}
\end{figure*}
\vspace{0.5cm}

\textbf{Query Types:}
    \begin{itemize}
    \setlength{\itemsep}{1pt}
    \setlength{\parskip}{1pt}
    \setlength{\parsep}{0pt}
        \item QS (Structured Query): Structured command format, e.g. \textit{open(microwave)}.
        \item QU (Unstructured Query): Natural language command format, e.g. \textit{``Can you please open the microwave?"}.
    \end{itemize}

    \textbf{Query Abstraction: }
    \begin{itemize}
    \setlength{\itemsep}{1pt}
    \setlength{\parskip}{1pt}
    \setlength{\parsep}{0pt}
        \item AS (Abstract Specific): Directly and clearly specified arguments, e.g. \textit{``Pick the banana from the fridge."}.
        \item AN (Abstract by Name): An argument is referred to by category or general attribute, e.g. \textit{``Can you turn off the electronic device on the shelf?"}.
        \item AR (Abstract by Spatial Relationship): An argument is specified through its spatial relation to another item, e.g. \textit{``Please grab the utensil to the left of the microwave"}.
        \item AC (Abstract by Conditioned State): An argument is specified through its conditioned state, e.g. \textit{``Please pick the clean utensil from the counter."}.
    \end{itemize}

\begin{figure}[!h]
\begin{center}
\includegraphics[width=0.65\columnwidth]{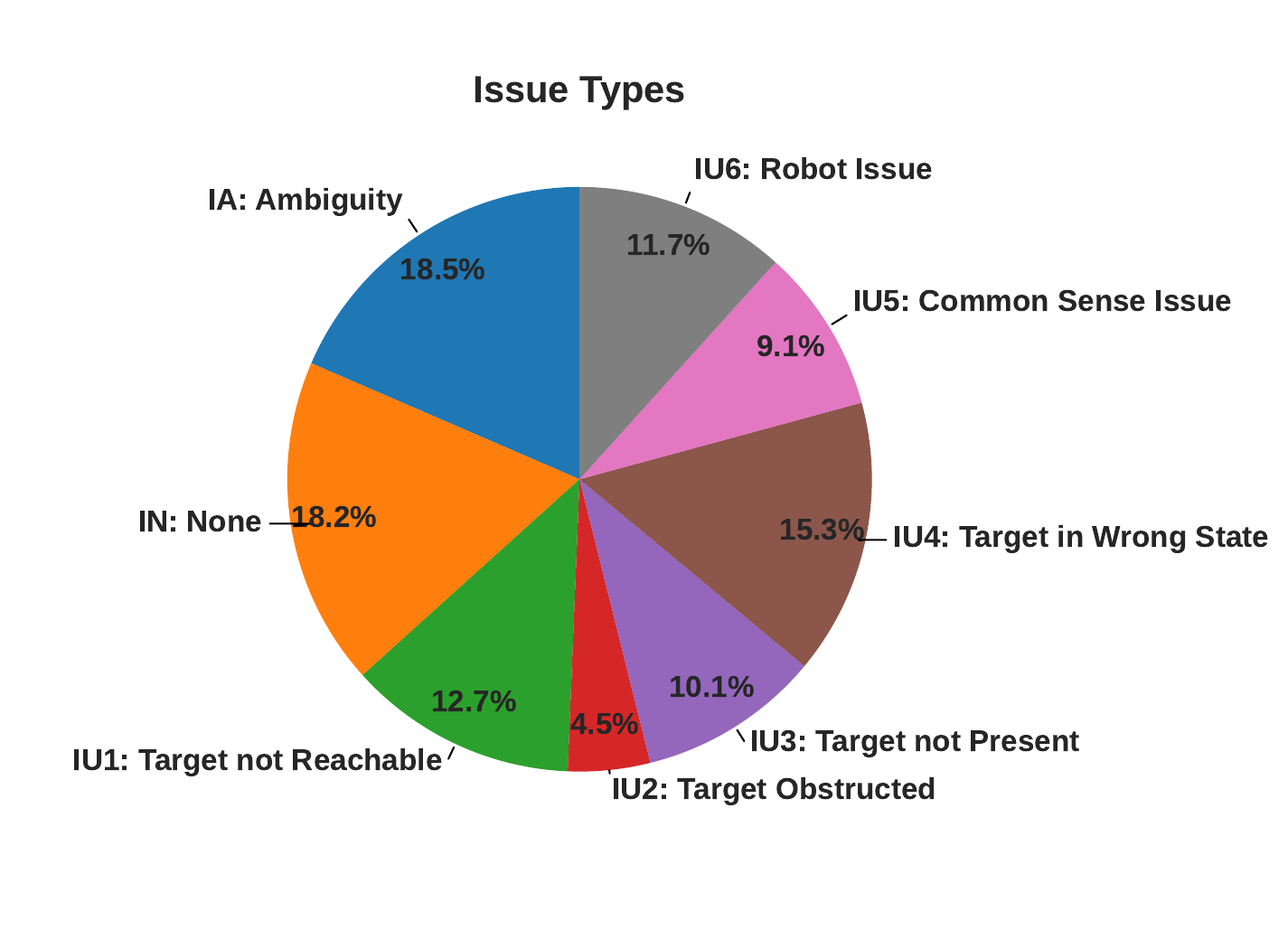}
\end{center}
\vspace{-1cm}
\caption{Dataset distribution by associated issue type.}\label{fig:issuetypes}
\end{figure}

\textbf{Issue Types:} As illustrated in Figure~\ref{fig:issuetypes}, the test cases are divided into different issue types to be detected, as:
    \begin{itemize}
    \setlength{\itemsep}{1pt}
    \setlength{\parskip}{1pt}
    \setlength{\parsep}{0pt}
        \item IA (Ambiguity): Two or more items match the query.
        \item IU (Unfeasibility):
        \begin{itemize}
        \setlength{\itemsep}{1pt}
        \setlength{\parskip}{1pt}
        \setlength{\parsep}{0pt}
        \item IU1: Target not within reach (too far).
        \item IU2: Target obstructed.
        \item IU3: Target not present.
        \item IU4: Target in wrong state e.g., already off.
        \item IU5: Common sense, e.g. target with wrong properties: not openable.
        \item IU6: Robot issue e.g., holding/not holding an object.
        \end{itemize}
        \item IN (No issue): No issues detected.
    \end{itemize}

\subsubsection{Tool Implementation}\label{sec:tool_impl}
The tool implementation in our evaluation is targeted for the described dataset, composed of simulated AI2THOR-generated events. This implementation utilizes RGB-D observations, for which object instance masks and environment state metadata have been generated in simulations. The list of objects is generated from the detected instance masks, with the objects' states and properties detailed in the accompanying metadata. The distance to each object is determined using the depth mask. To assess object relationships, we adapt the heuristics outlined in REFLECT~\cite{liu2023reflect}. Instance masks and the observed depth are used to generate 3D object semantic point clouds and bounding boxes, enabling the computation of eight spatial relationships between objects: inside, on top of, to the left of, to the right of, above, below, occluding, and near. For the example experiments using real-world images (see Figure~\ref{fig:example_real} in the appendix), the objects' bounding boxes are obtained through OWL-ViT~\cite{minderer_simple_2022} object detection model, and used as input prompts to obtain mask detections through the Segment Anything Model (SAM)~\cite{kirillov_segment_2023}. For state and property detection, we adopt the methodology described in~\cite{liu2023reflect}, computing the cosine similarity of the CLIP embedding~\cite{radford_learning_2021} between the object's image and a predefined list of object state labels. Finally, even though the agent is prompted to call tools with arguments that match detected object names, fuzzy string matching is implemented in both simulation and real-world versions to accommodate arguments that do not exactly match the names of the objects detected in the scene.
Note that these methods can be adapted or substituted with more advanced techniques~\cite{kenghagho_kenfack_robotvqa_2020,migimatsu2022grpred,santoro2017relreasoning}, by updating the tool implementation. We recognize that our system depends on the successful application of these tools, similar to the methods under comparison. However, creating a more advanced version of these tools falls outside the scope of this study, which concentrates on the components of the agent framework outlined in the methodology.
\subsection{Issue Detection and Explanation Evaluation}\label{sec:eval_issuedetect}
\subsubsection{Method Comparison}\label{sec:evalmethods}
To showcase the benefits of our framework regarding issue detection and explanation, we compare with the following baseline methods:

\begin{itemize}
    \item Action Model Precondition Check (Precond): checking predefined preconditions for each action, e.g., \textit{place(object, location)} includes \textit{robot\_holding(object)} as one of its preconditions. See~\ref{app2_baseline_precond} for details on these checks and their use for issue detection and explanation generation. Unstructured or abstract queries are not supported by this method.
    \item Visual Question Answering (VQA): asking LLaVA-NeXT~\cite{liu2024llavanext}, a state-of-the-art visual question answering model, for a potential issue in executing the action based on the image provided. The prompt provided to the model can be found in~\ref{app2_baseline_vqa}.
    \item LLM with Visual Observation (LLMVisualObs): providing the LLM with our system prompt, replacing the tool related sections with a visual observation obtained from the full scene graph as implemented in~\cite{liu2023reflect}. The prompt injected into the LLM can be found in~\ref{app2_baseline_visualobs}.
    \item LLM agent with VQA (LLMAgentVQA): our approach, replacing the tool implementations by the VQA model LLaVA-NexT~\cite{liu2024llavanext}. See~\ref{app2_baseline_llmvqa} for the prompt provided to the VQA model.
\end{itemize}

\begin{figure}
\begin{center}
\includegraphics[width=0.6\columnwidth]{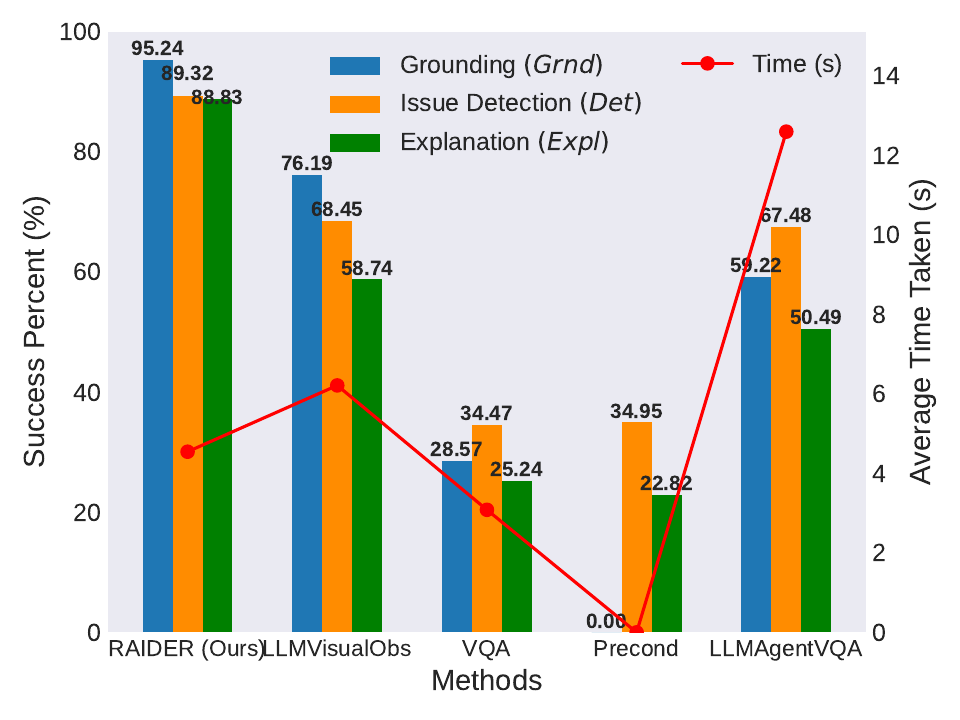}
\end{center}
\vspace{-0.4cm}
\caption{Result comparison between our method and the baseline approaches for grounding, issue detection and explanation for all the evaluated test cases. RAIDER outperforms other methods across all metrics, with an average processing time of 4.54 seconds.}\label{fig:results_methods}
\end{figure}

\textbf{Method Comparison Results.}
Figure~\ref{fig:results_methods} presents a result comparison between our method and the baseline approaches. Overall, RAIDER outperforms other methods across all metrics: grounding, issue detection, and explanation, achieving success rates of 95.23\%, 89.32\% and 88.83\%, respectively. LLMVisualObs ranks second, while LLMAgentVQA, VQA and Precond show significantly lower success metrics. These methods struggle to achieve grounding, not handling abstract queries effectively, and support only limited checks, reducing the success in issue detection and explanation. The high success rate of explanations in our system (88.83\%) is crucial, especially when compared to the next highest achieving method, LLMVisualObs, which achieves 76.19\%. Successfully explaining the issue demonstrates the success of the full pipeline, including grounding and issue detection, and facilitates effective recovery from identified issues. RAIDER maintains an average processing time of 4.54 seconds, which is shorter than that of LLMVisualObs and LLMAgentVQA, despite being longer than VQA and Precond, which show much lower success rates. This highlights a critical trade-off between speed and effectiveness.

\textbf{Method Comparison by Issue Type.} The results presented in Table~\ref{tab:expl-issue} provide a comparative analysis of our method against the baseline methods, based on their ability to detect different types of issues. As described in Section~\ref{sec:dataset}, the issue types include ambiguity (IA), various unfeasibility categories (IU1 through IU6), as well as no detected issues (IN). The metric \textit{Expl} (explanation success) is taken as the success metric, indicating that both grounding (if required) and issue detection have been correctly achieved beforehand.

RAIDER demonstrates superior performance across all issue types, particularly in ambiguity detection IA, no issues IN, unreachable targets IU1, non-existent targets IU3 and wrong target states IU4. Our approach has more difficulty in obstruction detection IU2 and wrong object properties detection IU5, implying that the system occasionally fails to check for an obstruction or for an object property, which is sometimes assumed by the agent.

Regarding the comparing methods, LLMVisualObs stands out in detecting ambiguities and object states (IU4) due to its comprehensive object listings but struggles with the remaining issues, meaning that the LLM fails to consider important information when the full scene description is provided. Furthermore, unreachable targets IU1 are not supported as they are not provided in the description. Precond performs well in detecting robot-related issues IU6 and no issues, thanks to its predefined checks, but fails in ambiguity detection and unreachable targets IU1, reflecting its limitations with abstract queries and spatial reasoning. VQA is effective in affirming no issues but struggles with detecting existent issues, indicating its limitations in handling complex scenarios. LLMAgentVQA shows better performance by leveraging the contextual understanding of the LLM and visual capabilities of the VQA model, but accurately identifies less than half of the issues for most issue types. 

RAIDER's overall high performance across all issue types demonstrates the effectiveness of its combined tool-based approach and reasoning mechanism for detecting a number of different issues, offering a significant improvement over existing methods.

\begin{table}
\centering
\resizebox{0.6\columnwidth}{!}{%
\begin{tabular}{@{}lllllllll@{}}
\toprule
\textbf{}            & \textbf{IA}  & \textbf{IN}   & \textbf{IU1}  & \textbf{IU2} & \textbf{IU3}  & \textbf{IU4}  & \textbf{IU5}  & \textbf{IU6}  \\ \midrule
\textbf{LLMVisualObs} & 83.3 & 66.7 & 0    & 58.3 & 63.6 & 87.1 & 42.1 & 40.7 \\
\textbf{Precond}      & 0    & 46.2 & 0 & 0    & 13.6 & 16.1 & 26.3 & 63   \\
\textbf{VQA}          & 8.3  & 69.2 & 10.7 & 0    & 36.4 & 12.9 & 31.6 & 11.1 \\
\textbf{LLMAgentVQA}  & 66.7 & 38.5 & 46.4 & 8.3  & 77.3 & 48.4 & 47.4 & 63   \\ \midrule
\textbf{RAIDER (Ours)} & \textbf{100} & \textbf{92.3} & \textbf{85.7} & \textbf{75}  & \textbf{90.9} & \textbf{96.8} & \textbf{68.4} & \textbf{85.2} \\ \bottomrule
\end{tabular}%
}
\caption{Explanation Success by Issue Type. RAIDER outperforms comparing methods across all issue types.}
\label{tab:expl-issue}
\end{table}


\subsubsection{Prompt Ablations}\label{sec:evalpromptabl}

In this section, we analyze the performance of our proposed prompt (QGEN\_GRND) against three ablations: BASIC, QGEN, and GRND. The ablations consist in modifying the \textit{Procedure Steps} in the prompt outlined in Figure~\ref{fig:prompt_templ} as follows:

\begin{itemize}
    \item QGEN\_GRND (Ours): our prompt, as described in Section~\ref{sec:prompt}.
    \item BASIC: removing the object grounding and the question generation in the procedure steps (directly generating tool calls). Note that this represents a regular REACT~\cite{yao2023react} ``think, act, observe" cycle.
    \item QGEN: maintaining the question generation step and removing the object grounding step.
    \item GRND: maintaining the grounding step and removing the question generation step.
\end{itemize}
\begin{figure}[hb]
\begin{center}
\includegraphics[width=0.6\columnwidth]{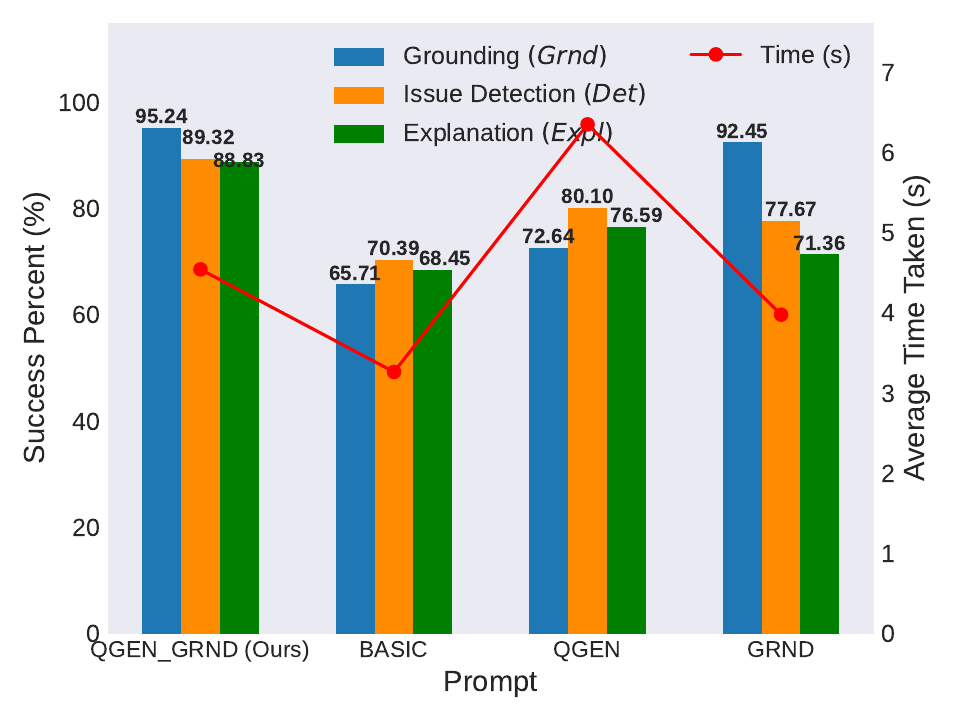}
\end{center}
\vspace{-0.4cm}
\caption{Result comparison for different prompt ablations across all test cases. The superior success rates obtained with QGEN\_GRND (RAIDER's prompt containing both grounding and question generation steps) demonstrate the importance and efficacy of every step in the procedure.}\label{fig:results_prompt}
\end{figure}
\textbf{Prompt Ablation Results.} The results of the prompt ablation study, as illustrated in Figure~\ref{fig:results_prompt}, demonstrate that the full prompt (QGEN\_GRND), which includes both object grounding and question generation steps, effectively achieves the highest success rates across all metrics. The BASIC prompt, which omits both grounding and question generation steps, performs the worst across all metrics, with a success rate reduction of 29.5\% in grounding, 18.9\% in issue detection, and 20.4\% in issue explanation. Removing only the grounding step in the QGEN prompt significantly lowers the grounding success rate by 22.6\%, affecting as well the issue detection and explanation success rates, which are lowered by 9.7\% and 12.2\% respectively. This indicates that the grounding stage is essential for the correct processing of the query. When removing the
question generation step in the GRND prompt, the issue detection and explanation success rates drop by 11.7\% and 17.5\%, respectively, while maintaining a high grounding success rate of 92.4\%. This highlights the crucial role of the question generation step in accurately detecting and explaining issues.

The time analysis indicates that the question generation step adds some processing time. In the QGEN prompt, the significant increase is due to the agent asking questions about queries that have not been properly grounded, resulting in less satisfactory tool returns and longer processing times. For the QGEN\_GRND prompt, although the processing time is slightly increased compared to the ablations not including question generation, it is reduced compared to QGEN, and a good performance is achieved, reflecting a trade-off between time taken and performance. These findings confirm that the structured procedure steps of grounding and question generation defined in this work are key to effectively processing queries and detecting and explaining issues in robotic action execution.
\begin{figure}
\begin{center}
\includegraphics[width=0.6\columnwidth]{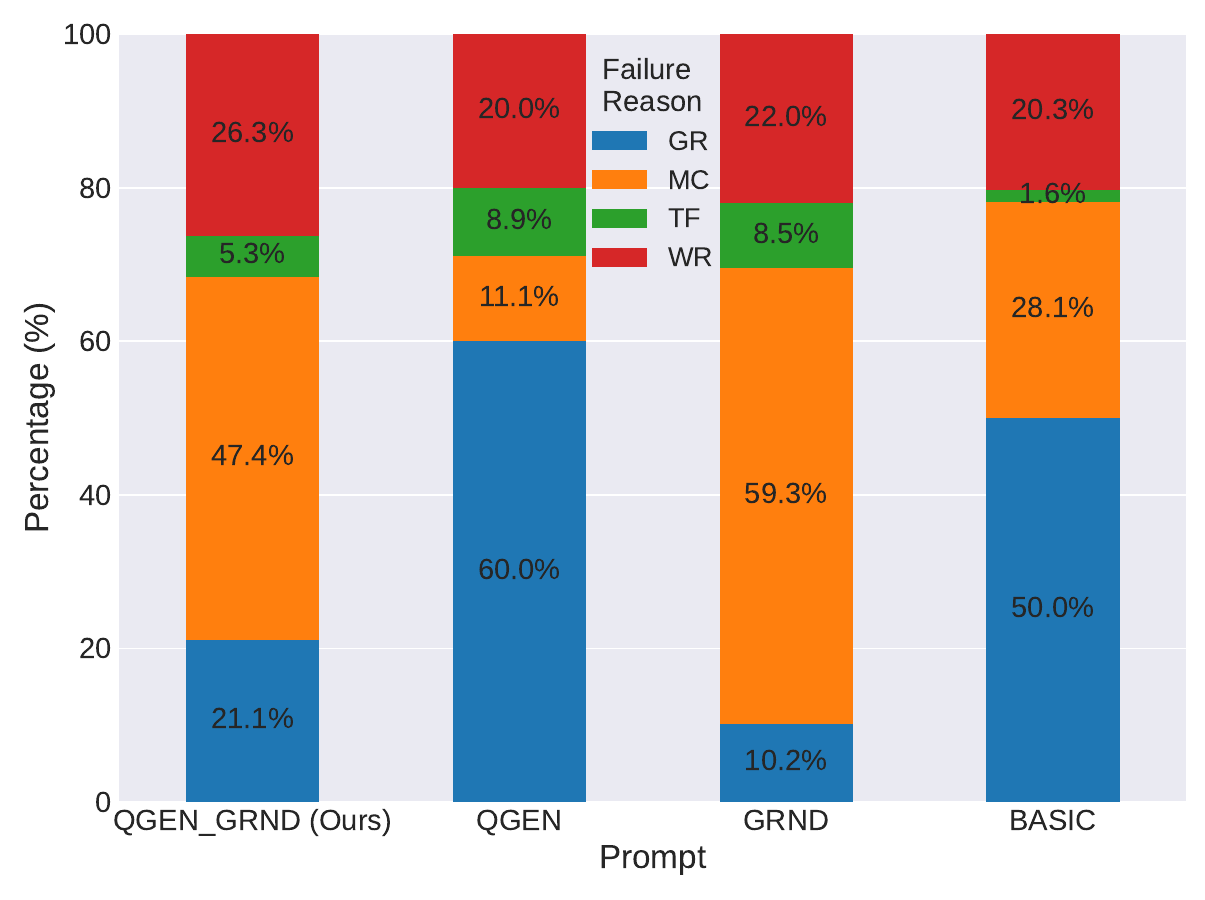}
\end{center}
\vspace{-0.5cm}
\caption{Distribution of failure reasons across failed test cases (where no correct issue explanation was provided) for our method and various prompt ablations.}\label{fig:results_prompt_failtypes}
\end{figure}

\subsubsection{Failure Reason Analysis}\label{sec:eval_failures}
We classify the system's failure to provide a correct explanation (measured by the metric \textit{Expl}) by the following reasons:
\begin{itemize}
    \setlength{\itemsep}{1pt}
    \setlength{\parskip}{1pt}
    \setlength{\parsep}{0pt}
    \item Grounding (GR): The query was not successfully grounded to specific objects in the scene, e.g., `the fruit in the fridge' was not grounded to `Apple'.
    \item Missed Check (MC): A necessary check during question or tool call generation was omitted, resulting in incomplete information, e.g., the object to pick is blocked by another object but this was not checked for.
    \item Wrong Reasoning (WR): Despite collecting all the required information, the system arrived at an incorrect conclusion, e.g., the system did not recognize that a chair cannot be picked up.
    \item Wrong Tool Call Format (WF): The tool call format was not respected, resulting in failed tool calls and missing information.
\end{itemize}
\textbf{RAIDER's Failure Reasons.} While our method achieves a high \textit{Expl} success rate of 88.83\%, it can still struggle with ensuring all necessary checks are performed (47.4\% of failures), and sometimes reaches incorrect conclusions despite having the required information (representing 26.3\% of the failures), as shown in Figure~\ref{fig:results_prompt_failtypes} for QGEN\_GRND. Grounding errors represent 21.1\% of failures, while wrong tool call format errors constitute the least common failure type (5.3\%). To further improve the system, particularly in addressing missed checks and wrong reasoning, it may be beneficial to inject additional task-specific knowledge into the prompt. This constitutes a trade-off between enhancing the system's robustness and maintaining its adaptability to diverse tasks.

\textbf{Prompt Ablations' Failure Reasons.} The failure reason distribution across different prompt ablations (Figure~\ref{fig:results_prompt_failtypes}) emphasizes again the critical roles of both object grounding and question generation for effective issue explanation. When using the QGEN and BASIC prompt (no grounding), over half of the failures (60.0\% and 50.0\%) are due to grounding errors, highlighting the importance of the grounding step. The GRND prompt's results show that question generation is crucial in achieving complete information gathering, with 59.3\% of failures attributed to missed checks.

\subsubsection{Repeatability}\label{sec:evalrep}
\begin{figure}
\begin{center}
\includegraphics[width=\columnwidth]{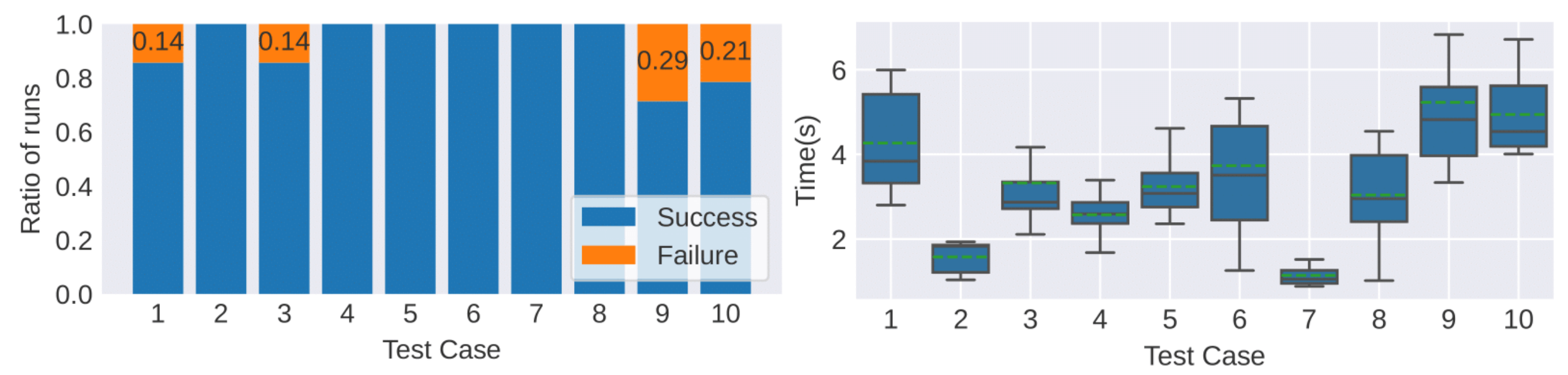}
\end{center}
\vspace{-0.4cm}
\caption{Issue explanation success and time repeatability across test cases, where each test case is run 10 times.}\label{fig:repeatability}
\end{figure}

The success and time repeatability of the system across test cases is examined as shown in Figure~\ref{fig:repeatability}. Ten test cases were randomly selected, and each test case was run 10 times, with the model's temperature parameter set to 0 to reduce randomness in the outputs.
The analysis shows that the overall success rate remains high across most test cases. Cases 9 and 10 show lower success rates, highlighting the potential variability introduced by the LLM. The higher associated execution times suggest a correlation between execution time and success rates, indicating that the LLM may be taking longer to generate responses due to looping and encountering difficulties in arriving at the correct answer. Overall, these findings indicate that while the system is fundamentally robust, there are opportunities for improvement in specific scenarios to address the variability arising from the LLM. Some techniques to mitigate this include improved  self-verification mechanisms and memory management.

\subsection{Interactive Action Recovery Evaluation}\label{sec:evalrecov}
In this section, we assess how RAIDER's issue explanations can be leveraged for generating recovery plans that might include interactive actions, such as asking for information from a human operator or communicating relevant details. Our evaluation is based on successful test cases derived from the dataset outlined in Section~\ref{sec:dataset}, for which issues were accurately identified and explained by RAIDER. As detailed in Section~\ref{sec:method_recovery}, our recovery action LLM ingests a query that contains the action the robot needs to recover from, the current object being held by the robot, and the identified issue along with its explanation. In Table~\ref{tab:eval_recov}, we compare the \textit{Recov\_Plan} success obtained from using our prompt and two other variations: providing the scene description (from the method LLMVisualObs described in Section~\ref{sec:evalmethods}) instead of RAIDER's issue explanation, and providing both the scene description and the issue explanation, as done in~\cite{liu2023reflect}.
The results showcase the importance of issue explanations in successfully generating recovery plans. When relying solely on scene descriptions, the overall success rate is significantly lower at 36.8\%. Combining scene descriptions with issue explanations results in an improved success rate of 69.03\%, but still underperforms compared to using explanations alone, which yields a \textit{Recov\_Plan} success rate of 85.1\%. This indicates that additional scene details may confuse the LLM with irrelevant information. Examining specific issue types, our recovery system achieves correct recovery actions in over 82\% of cases across all categories, except for obstruction issues, where success drops to 55.6\%. Interestingly, for obstruction issues, including the scene description appears beneficial. For the rest of issue types, providing the issue explanation on its own significantly enhances results. On average, the time taken to generate these plans is 1.27 seconds, demonstrating the system's applicability in real-time scenarios. Overall, these findings highlight that the explanations generated by RAIDER are sufficiently informative to assist an LLM in formulating effective recovery plans.

\begin{table}
\centering
\resizebox{0.6\columnwidth}{!}{%
\begin{tabular}{@{}l|l|lllllll@{}}
\toprule
\textbf{} &
  \textbf{All} &
  \textbf{IA} &
  \textbf{IU1} &
  \textbf{IU2} &
  \textbf{IU3} &
  \textbf{IU4} &
  \textbf{IU5} &
  \textbf{IU6} \\ \midrule
\textbf{Scene} &
  36.8 &
  81.1 &
  12.5 &
  33.3 &
  15.8 &
  33.3 &
  0.0 &
  34.8 \\
\textbf{\begin{tabular}[c]{@{}l@{}}Explanation and Scene\end{tabular}} &
  69.0 &
  81.1 &
  41.7 &
  \textbf{66.7} &
  68.4 &
  60.0 &
  91.3 &
  78.3 \\ \midrule
\textbf{Explanation} & \textbf{85.1} & \textbf{86.5} & \textbf{86.4} & 55.6 & \textbf{94.7} & \textbf{83.9} & \textbf{92.3} & \textbf{82.6} \\ \bottomrule
\end{tabular}%
}
\caption{Action Recovery Success (\textit{Recov\_Plan}) across Issue Types. Comparison between using a prompt with scene description, with scene description and RAIDER's issue explanation, and with RAIDER's issue explanation only.}
\label{tab:eval_recov}
\end{table}

\subsection{Key Findings}\label{sec:key_find}
This section summarizes the key findings of our research, highlighting both the strengths and limitations of the system.
\subsubsection{Strengths}
Combining grounded tools with LLMs for selective and contextual information gathering provides:
\begin{itemize}
\item \textbf{Increased robustness in issue detection and explanation:} With an \textit{Expl} success rate of 88.83\% (Figure~\ref{fig:results_methods}), RAIDER outperforms methods that utilize full scene descriptions, visual language models, or predefined precondition checks, across various issue types (Table~\ref{tab:expl-issue}).
\begin{figure*}[!b]
    \centering
    \includegraphics[width=\textwidth]{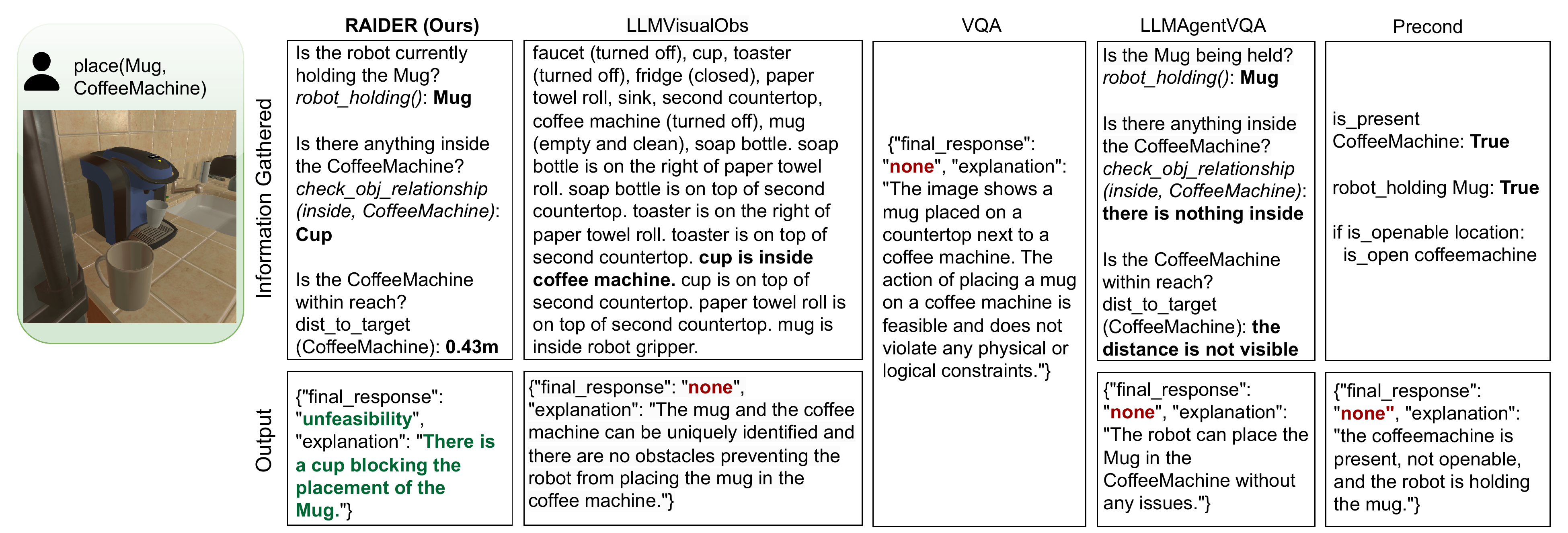}
    \caption{Comparing information gathering through different methods for issue detection. Our method actively gathers contextually relevant information, successfully detecting the issue. Other methods gather irrelevant, inaccurate, or insufficient information, reaching a wrong conclusion.}
    \label{fig:info_gathered_methods}
\end{figure*}

\item \textbf{Most accurate and useful information gathered:} As shown in Figure~\ref{fig:info_gathered_methods}, selectively querying tools based on context effectively identifies the issue, surpassing open-ended scene description methods that process the entire scene. These methods often result in insufficient or inaccurate information, as seen in the cases of VQA and LLMAgentVQA. In contrast, LLMVisualObs, which receives the complete scene graph description, struggles to focus on the relevant information, failing to detect the issue.

\item \textbf{Increased adaptability to different scenarios:} By dynamically generating precondition questions, RAIDER effectively addresses diverse issue scenarios, which models based on fixed action preconditions would fail to deal with. Taking the scenario in Figure~\ref{fig:info_gathered_methods}, where a mug needs to be placed in a coffee machine, most traditional models would overlook any items already inside since checking that condition is not part of their standard ``plac'' action model. Addressing this requirement would require defining a specific action in the model for such place location.

\item \textbf{Improved grounding of diverse levels of action abstractions:} As shown in Table~\ref{tab:expl-issue}, RAIDER can effectively ground a higher range of instructions, expressed in different types of abstraction where the targets are not explicitly named. Some examples can be found in \ref{app3_qualitative_examples}. 
\end{itemize}

\textbf{Grounding and question generation steps have critical roles.} Each component of RAIDER's ``Ground, Ask\& Answer, Issue" procedure steps is crucial, highlighted by a drop in success rates of around 20\% when either component is removed (Figure~\ref{fig:results_prompt}). As shown in Figure~\ref{fig:results_prompt_failtypes}, object grounding is essential for accurate query processing (majority of grounding errors when removed) while question generation is key for effective check generation (majority of missed checks errors when removed).

\textbf{Program flow manager warning messages help the LLM in self-reflection and tool call recovery.} Warning messages are effective at flagging LLM responses containing incorrect tool call names, arguments or formats, prompting the LLM to self-correct. In our findings, warnings were identified in 5.3\% of the test cases, and the LLM successfully corrected itself in 86.8\% of those instances. An illustration of the four warning checks established in the program can be found in Figure~\ref{fig:warnings}.

\textbf{Generated explanations support effective recovery.} Providing RAIDER's issue explanations significantly improves the success rate of recovery plans, achieving an 85.1\% success rate compared to 36.8\% when relying solely on scene descriptions (Table~\ref{tab:eval_recov}).

\textbf{Interactive Recovery is important for resolving issues.} Interactive recovery actions allow for a greater variety of situations to be addressed, such as ambiguity, missing objects, and common-sense problems (as illustrated in the examples in Figure~\ref{fig:llmagent_ex}), which would be difficult or not possible to resolve without interactive actions. 

\textbf{RAIDER's modular architecture allows for straightforward adaptation to various scenarios and issue types.} The structured framework and modular prompt design allows for easy expansion of the agent’s objectives and capabilities by integrating new tools, providing a versatile solution that can address a wide range of challenges in robotics, as demonstrated in the subsequent Section~\ref{sec:extend}.

\subsubsection{Limitations}
We identify a number of limitations, presenting potential opportunities for future work:

\textbf{Robustness versus adaptability trade-off.} The failure reasons (Figure~\ref{fig:results_prompt_failtypes}) show that despite RAIDER's high success rate, its performance is occasionally affected by missed checks and incorrect reasoning. Addressing these issues by incorporating additional task-specific knowledge into the prompt can enhance robustness, with the potential cost of reduced adaptability to diverse tasks.

\textbf{LLM-induced variability impacts repeatability.} The variability of RAIDER's performance in some cases (Figure~\ref{fig:repeatability}) shows how LLMs may produce different responses to the same input due to their probabilistic nature, affecting repeatability and robustness. Some areas for improvement include further developing self-verification mechanisms and focused context management, rather than inputting the entire unfiltered conversation to the LLM.

\textbf{Reliance on accurate tool returns.} RAIDER's scene detection tools effectively address the cases discussed in this work, but would require enhancements for more complex scenarios. As suggested in \cite{liu2023reflect}, training larger models on robot data for object states, properties, and spatial relationships could improve accuracy \cite{li_embodied_2022, liu_visual_2023, kurenkov_modeling_2023}. However, note that this limitation is also present in the comparing methods.

\textbf{Low-level control.} RAIDER detects issues through high-level semantic precondition checks, assuming low-level checks such as calculating manipulation trajectories are integrated in the high-level skill execution. Although many studies \cite{ahn2022saycan, chen_open-vocabulary_2023, huang2023innermonologue} rely on this assumption, our system could be extended by integrating tools to compute essential low-level state information or constraints, such as grasping trajectories.

\section{Exploring RAIDER's Extensibility}\label{sec:extend}
We aim to showcase the extensibility of RAIDER in addressing a variety of scenarios, instructions, and issues beyond those assessed in our evaluation. The modular architecture of the system enables easy adaptation to different contexts by allowing changes to task objectives, toolsets, and scenario constraints. 
\subsection{Handling Human Preferences through Tool Addition}
\begin{figure*}[t]
    \centering
    \includegraphics[width=1\textwidth]{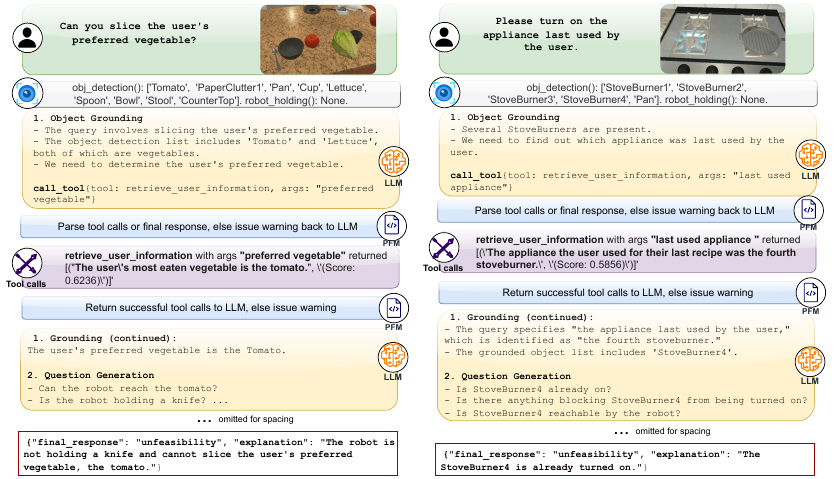}
    \caption{Handling user preferences through tool addition. By incorporating a tool to retrieve user information, RAIDER's capabilities are extended to deal with instructions incorporating user preferences and user activity history information, without additional system modifications.}
    \label{fig:user_prefs}
\end{figure*}
The first application demonstrates how RAIDER's toolset can be extended to handle user preferences within instructions by simply integrating a new tool. We introduce the tool \textit{retrieve\_user\_information(query)}, which semantically retrieves pertinent information from a corpus containing data on user preferences and activity history. The LLM formulates the query passed through the tool call, which returns the corpus entry with the highest cosine similarity to the embedded query. This process utilizes a sentence transformer model~\cite{reimers_sentence-bert_2019}. Note that this extension simply required the addition of the tool description to the list of tools in the prompt, and the incorporation of the tool's implementation in the code. The rest of the system remains unchanged, including the prompt, agent program flow, and LLM. We evaluate a number of instructions and present two examples in Figure~\ref{fig:user_prefs}, demonstrating the system's successful grounding of instructions that incorporate user preferences and activity history information.
The following corpus containing user information is provided to the tool \textit{retrieve\_user\_information(query)}:

\begin{footnotesize}
\begin{verbatim}
    "The user loves apples."
    "The user's most eaten vegetable is the tomato."
    "The user's favorite beverage is coke."
    "The user prefers potato chips as their snack."
    "The user's preferred book is `Don Quixote.'"
    "The user's most used cooking spice is oregano."
    "The user used a frying pan to cook their last meal."
    "The user was reading `1984' last night."
    "The appliance the user used for their last recipe was 
    the fourth stoveburner."
    "The most used stoveburner is the second one."
    "The user left the living room light on."
    "The most used light is the living room light."
    "The user enjoyed grapes at their last gathering."
    "The user bought a cutting board yesterday."
\end{verbatim}
\end{footnotesize}
This enhancement allows the system to consider individual user preferences when processing instructions, which would not be manageable by traditional comparative methods (such as VLMs, scene graphs, or predefined preconditions), as they lack the capability to support this type of data. By adding new tools, a diverse range of capabilities can be integrated into the system.

\subsection{Integrating RAIDER into Real-World Assistive Plans}\label{sec:extend_assist}
\begin{figure}[!h]
    \centering    \includegraphics[width=0.95\columnwidth]{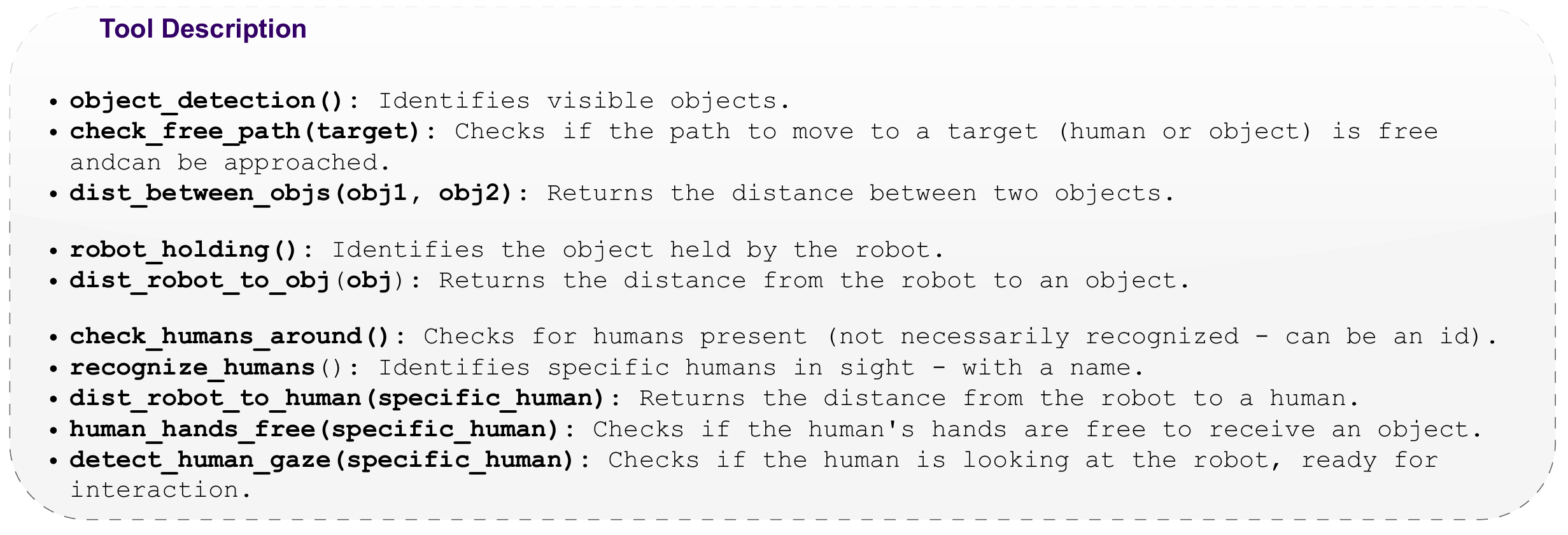}
    \caption{Modified toolset to deal with human assistive action issue detection.}
    \label{fig:demo_tool_list}
\end{figure}

The ultimate goal is to integrate RAIDER into real-world robotic scenarios, where robots typically execute instructions by decomposing them into a sequence of planned actions. This application demonstrates how RAIDER can be adapted to detect issues in human-assistive tasks within such structured plans. To achieve this, we modify the toolset to align with such scenario, by including human-related, robotic, and environmental tools, as depicted in Figure~\ref{fig:demo_tool_list}. Additionally, we introduce two minor modifications to the system prompt: specifying to check for both human and environmental issues and adding a constraint that requires handovers to occur within a distance of 0.5 meters. Apart from these adjustments, the core components of the system—including most of the prompt, the program flow, and the LLM—remain unchanged.

\begin{figure}[!h]
    \centering
    \includegraphics[width=\columnwidth]{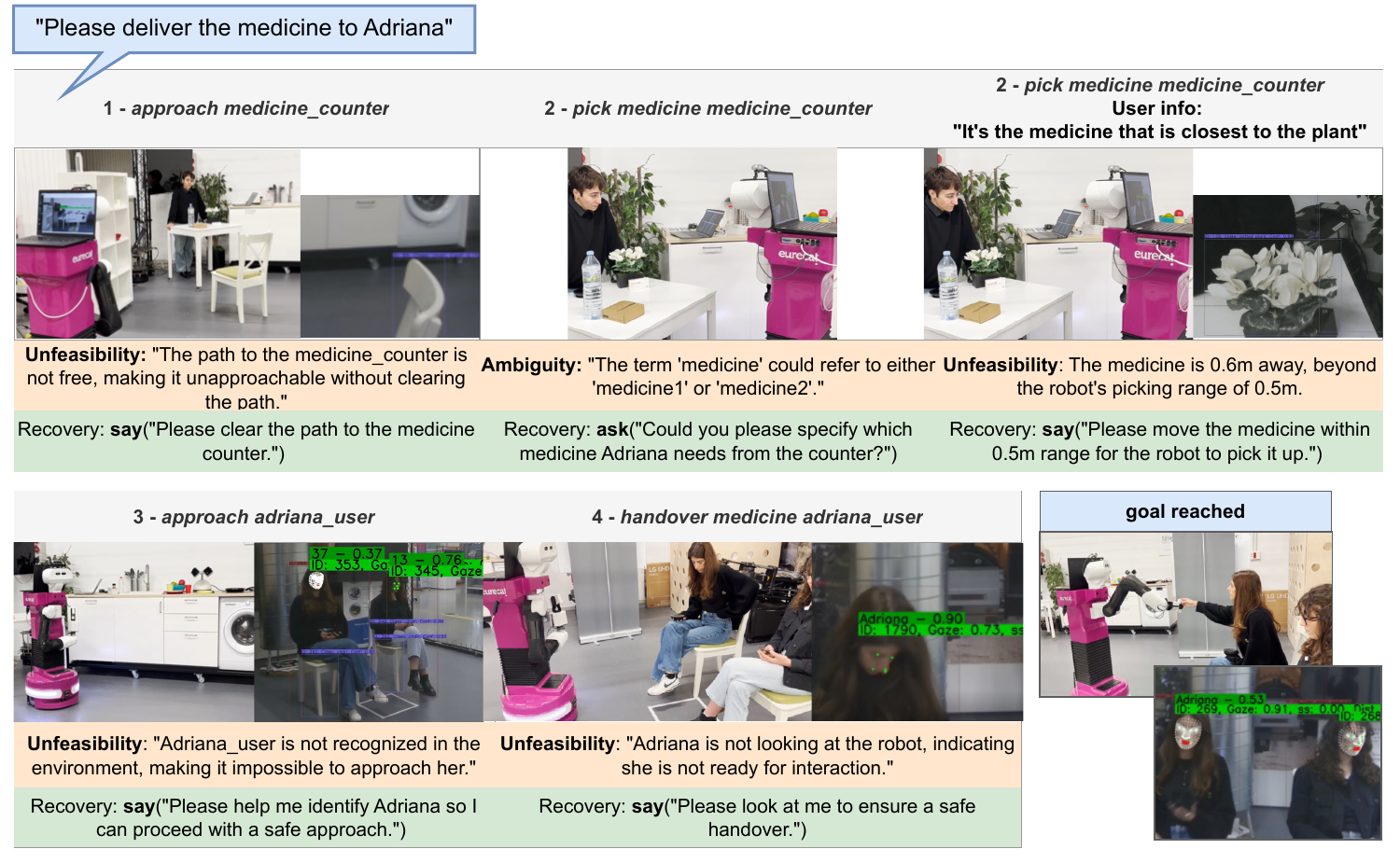}
    \caption{Extending RAIDER for assistive scenarios through task objective and toolset modification. This example illustrates a structured robotic plan for delivering a medicine to a user, highlighting detected issues at each step and corresponding recovery actions to ensure successful task completion.}
    \label{fig:demo}
\end{figure}

In this scenario, we test the instruction of delivering a medicine to a patient in the room. This high-level goal is translated into a PDDL plan following the methodology presented in \cite{izquierdo2024plancollabnl}. During execution, RAIDER is triggered at each PDDL action step, ensuring that no issues remain before proceeding. The plan consists of four actions: approaching the counter, picking up the medicine, approaching the user, and handing over the medicine. Figure~\ref{fig:demo} illustrates the sequence of actions, along with the detected issues and their corresponding recovery plans. Initially, the agent detects a blockage preventing the approach action, which is resolved when the human moves the chair. Next, an ambiguity issue arises as multiple medicines are present. The robot requests clarification as a recovery strategy, then uses the human's response to disambiguate and reassess feasibility. It then determines that the medicine is out of reach and asks the human to place it closer. Once the medicine is picked up, the robot proceeds to approach the user. Here, it fails to recognize them and communicates the issue. The user identifies themselves, allowing the robot to continue. Finally, during the handover, the agent detects that the patient is not looking, prompting it to request their attention. With this final issue resolved, the handover is successfully completed. The full RAIDER outputs containing the regulated flow between the LLM and the tools are provided in \ref{app5_demo}.

This showcases how, with minimal modifications to the prompt and the integration of new tools, the system can be effortlessly adapted to diverse scenarios while preserving its core architecture. Additionally, it highlights RAIDER’s capability to function within a real-world robotic plan, dynamically identifying, explaining, and resolving issues until the goal is successfully achieved. 

\section{Conclusions}\label{sec:conclusions}
We present RAIDER, an LLM-based system with access to grounded tools to detect and explain issues impeding the execution of robotic actions, expressed at various levels of abstraction. By employing a defined protocol,\emph{``Ground, Ask\&Answer, Issue"}, the framework dynamically generates precondition questions based on the context and invokes relevant tools to resolve them. The designed agent flow fosters an effective interaction between the LLM and the tools, improving the system's robustness and transparency via self-correction mechanisms that address the inherent LLM's limitations. Through selective and contextual reasoning and information gathering, our system demonstrates superior adaptability compared to traditional methods that rely on extensive action modeling, improved robustness over standalone visual language models, and enhanced efficiency by focusing on pertinent information rather than full scene graphs. RAIDER's modular architecture allows for easy expansion of the agent's capabilities through the integration of new tools, as shown in our evaluation, making it suitable for diverse issue detection scenarios in robotics. Furthermore, we show how the generated explanations serve as a foundation for creating recovery plans, that can involve interactive actions with human users. Overall, our findings underscore RAIDER's potential to enhance robot performance in dynamic human-centered environments by offering a structured and adaptable approach to issue detection and resolution. Future work will focus on addressing the identified limitations, particularly enhancing the system's robustness while maintaining its adaptability, along with further evaluation in real-world settings.
\clearpage


\bibliographystyle{elsarticle-num} 
\bibliography{references.bib, references_RAIDER.bib}

\clearpage

\appendix

\section{Qualitative Examples}\label{app3_qualitative_examples}
\begin{figure}[!h]
    \centering
\includegraphics[width=0.96\columnwidth]{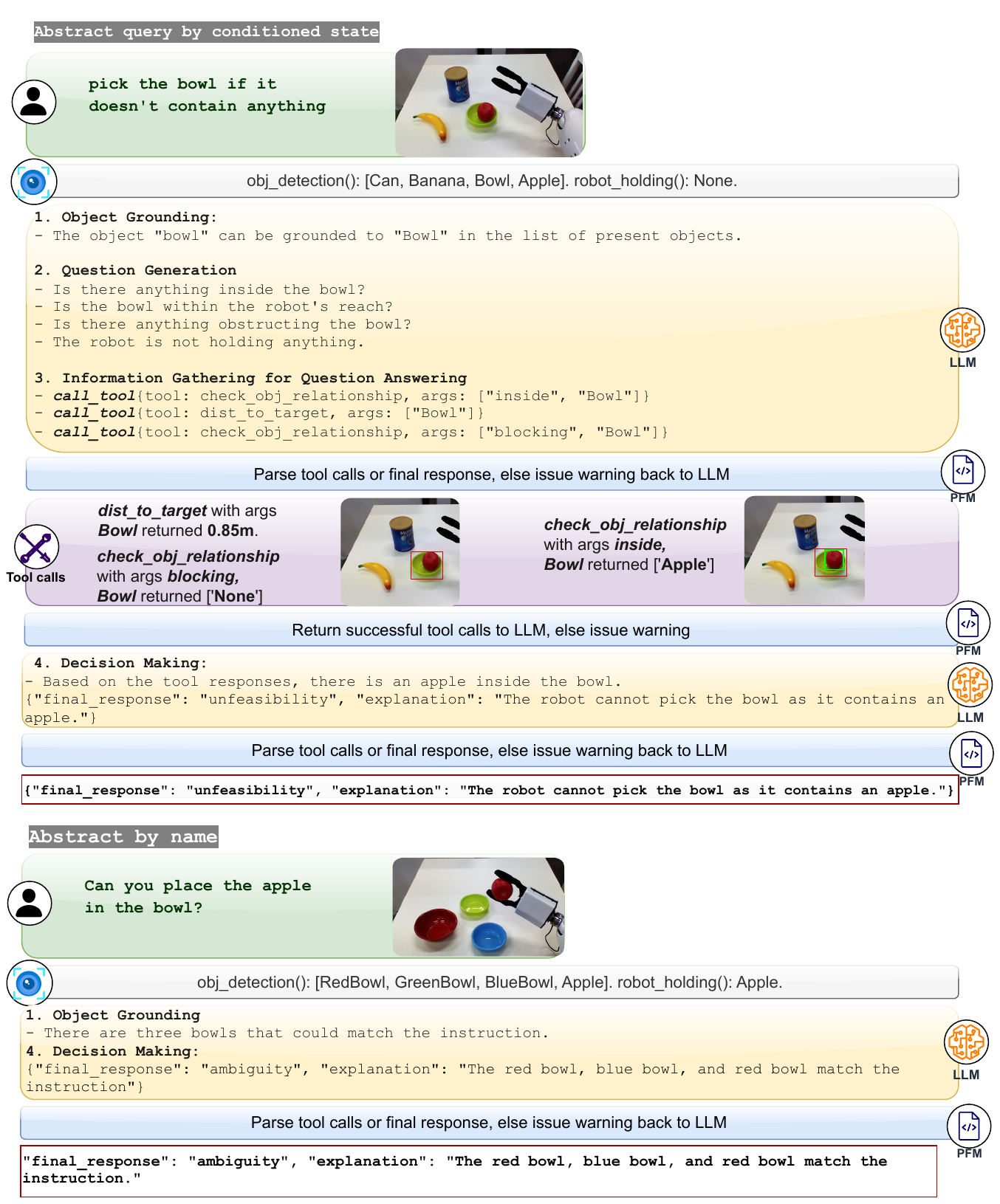}
    \caption{Qualitative example for query type AS (Abstract by Spatial relationship).}
    \label{fig:example_spatial_rel}
\end{figure}
\begin{figure}[!h]
    \centering
\includegraphics[width=0.96\columnwidth]{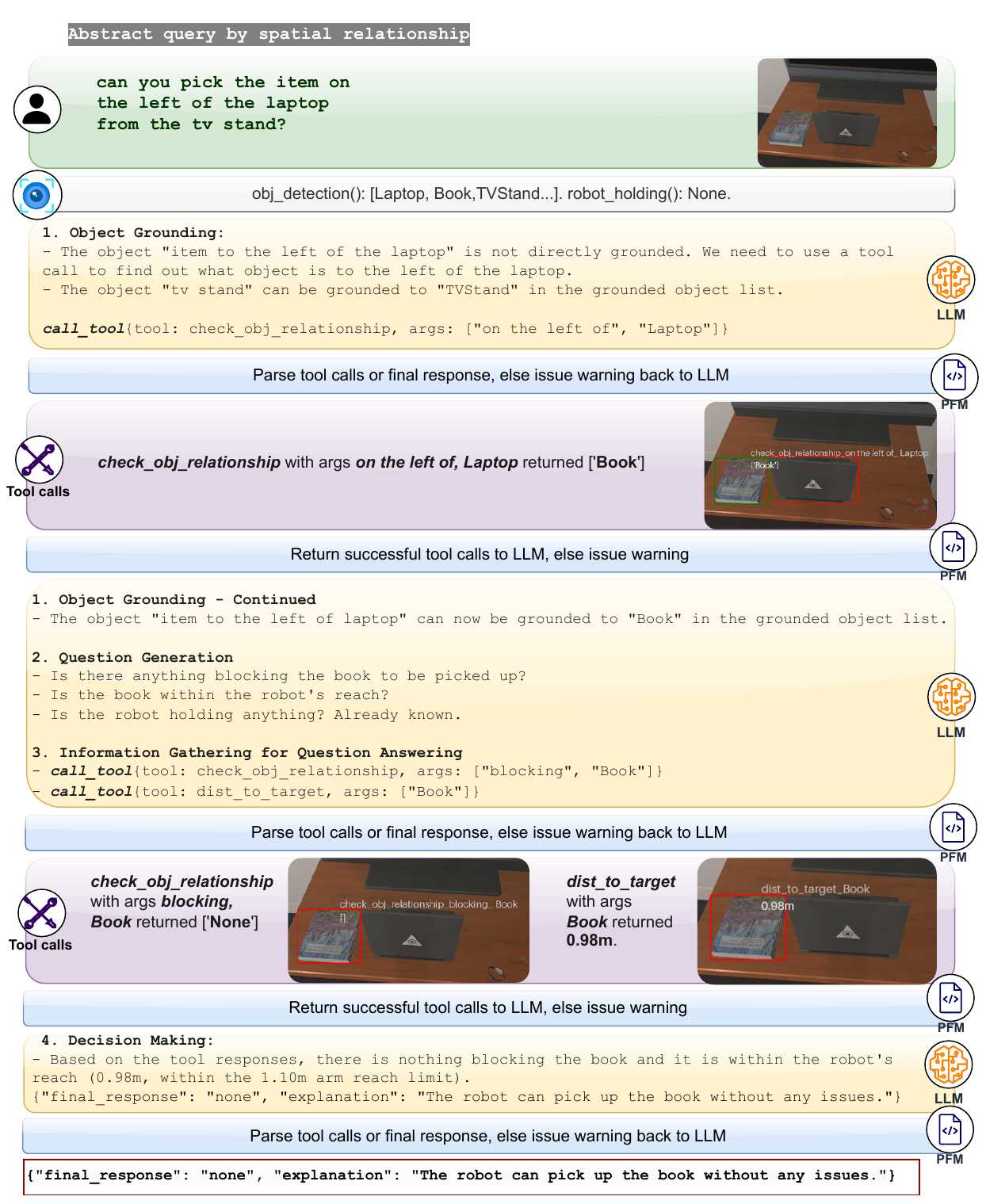}
    \caption{Qualitative examples for query types AC (Abstract by Conditioned state) at the top, and AN (Abstract by Name) at the bottom.}
    \label{fig:example_real}
\end{figure}

\section{Implementation Details of Baseline Methods}
\label{app2_baseline}
This section provides additional details for the implementation of the baseline methods described in Section~\ref{sec:evalmethods}.
\subsection{Action Model Precondition Check (Precond)}\label{app2_baseline_precond}
Table~\ref{tab:precond-checks} outlines the predefined checks performed during issue detection using the action model precondition check method. The implementation consists of nested precondition checks, which are manually translated to natural language for the issue explanation generation. These explanations are generated by either returning the corresponding negative check, or returning the appended positive checks if no unfeasibility issue is found.
\begin{table}[!h]
\centering
\resizebox{0.5\columnwidth}{!}{%
\begin{tabular}{@{}ll@{}}
\toprule
action(args)                            & Preconditions                               \\ \midrule
\multirow{3}{*}{pick(object, location)} & is\_present object                          \\
                                        & robot\_holding none                         \\
                                        & if is\_openable location: is\_open location \\ \midrule
place(object, location)                 & is\_present object                          \\
                                        & robot\_holding object                       \\
                                        & if is\_openable location: is\_open location \\ \midrule
\multirow{4}{*}{open(receptacle)}       & is\_present receptacle                      \\
                                        & robot\_holding none                         \\
                                        & is\_openable receptacle                       \\
                                        & is\_closed receptacle                     \\ \midrule
close(receptacle)                       & is\_present receptacle                      \\
                                        & robot\_holding none                         \\
                                        & is\_openable receptacle                         \\
                                        & is\_open receptacle                     \\ \midrule
turnon(object)                          & is\_present object                          \\
                                        & robot\_holding none                         \\
                                        & is\_toggeable object                        \\
                                        & is\_turned\_off object                      \\ \midrule
turnoff(object)                         & is\_present object                          \\
                                        & robot\_holding none                         \\
                                        & is\_toggeable object                        \\
                                        & is\_turned\_on object                       \\ \midrule
slice(object)                           & is\_present object                          \\
                                        & robot\_holding knife                        \\
                                        & is\_sliceable object                        \\
                                        & not\_sliced object                          \\ \bottomrule
\end{tabular}%
}
\caption{Action Precondition Checks}
\label{tab:precond-checks}
\end{table}

\subsection{Visual Question Answering (VQA)}\label{app2_baseline_vqa}
\begin{figure}[!h]
    \centering
    \includegraphics[width=0.92\columnwidth]{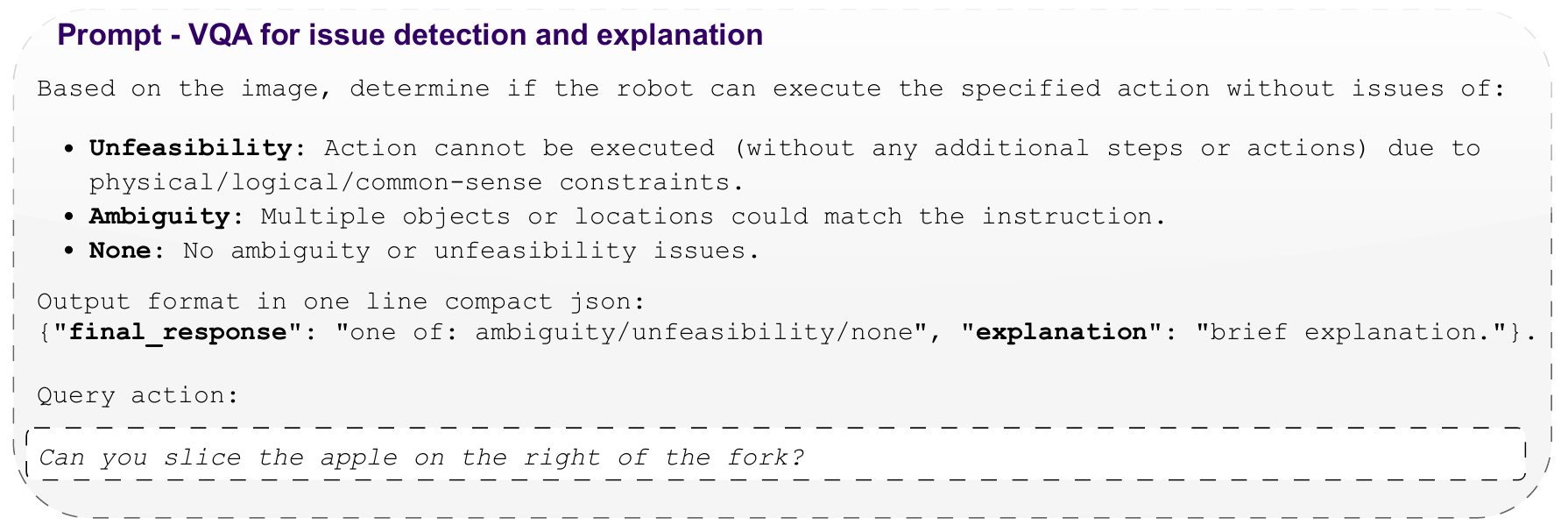}
    \caption{Prompt provided to the VQA model for issue detection and explanation, including an example query action.}
    \label{fig:prompt_vqa}
\end{figure}

The prompt provided to the VQA model LLaVA-NeXT~\cite{liu2024llavanext} for issue detection and explanation is provided in Figure~\ref{fig:prompt_vqa}.

\subsection{LLM with Visual Observation (LLMVisualObs)}\label{app2_baseline_visualobs}
\begin{figure}[!h]
    \centering
    \includegraphics[width=0.9\columnwidth]{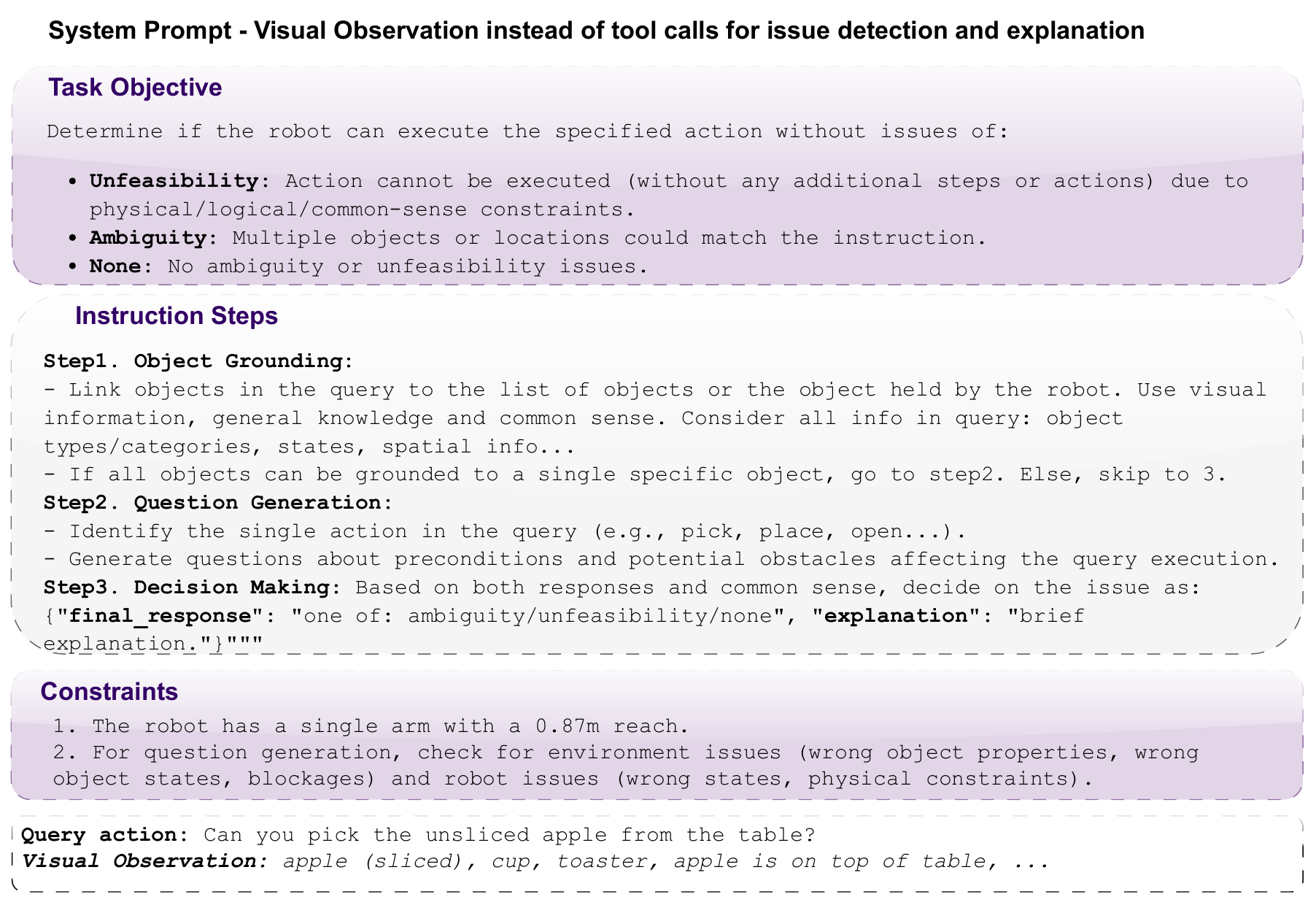}
    \caption{Prompt provided to the LLM for issue detection and explanation using the full scene visual observation. RAIDER's prompt is modified by removing the parts related to the tools, and the visual observation is provided instead.}
    \label{fig:prompt_visualobs}
\end{figure}

Figure~\ref{fig:prompt_visualobs} shows the prompt given to the LLM when replacing the tool calls by a single visual observation generated from a scene graph. The prompt structure is taken from our agent, removing any segments that refer to tool calls, and providing the full scene visual observation instead.
\subsection{LLM agent with VQA (LLMAgentVQA)}\label{app2_baseline_llmvqa}
\begin{figure}[!h]
    \centering
    \includegraphics[width=0.9\columnwidth]{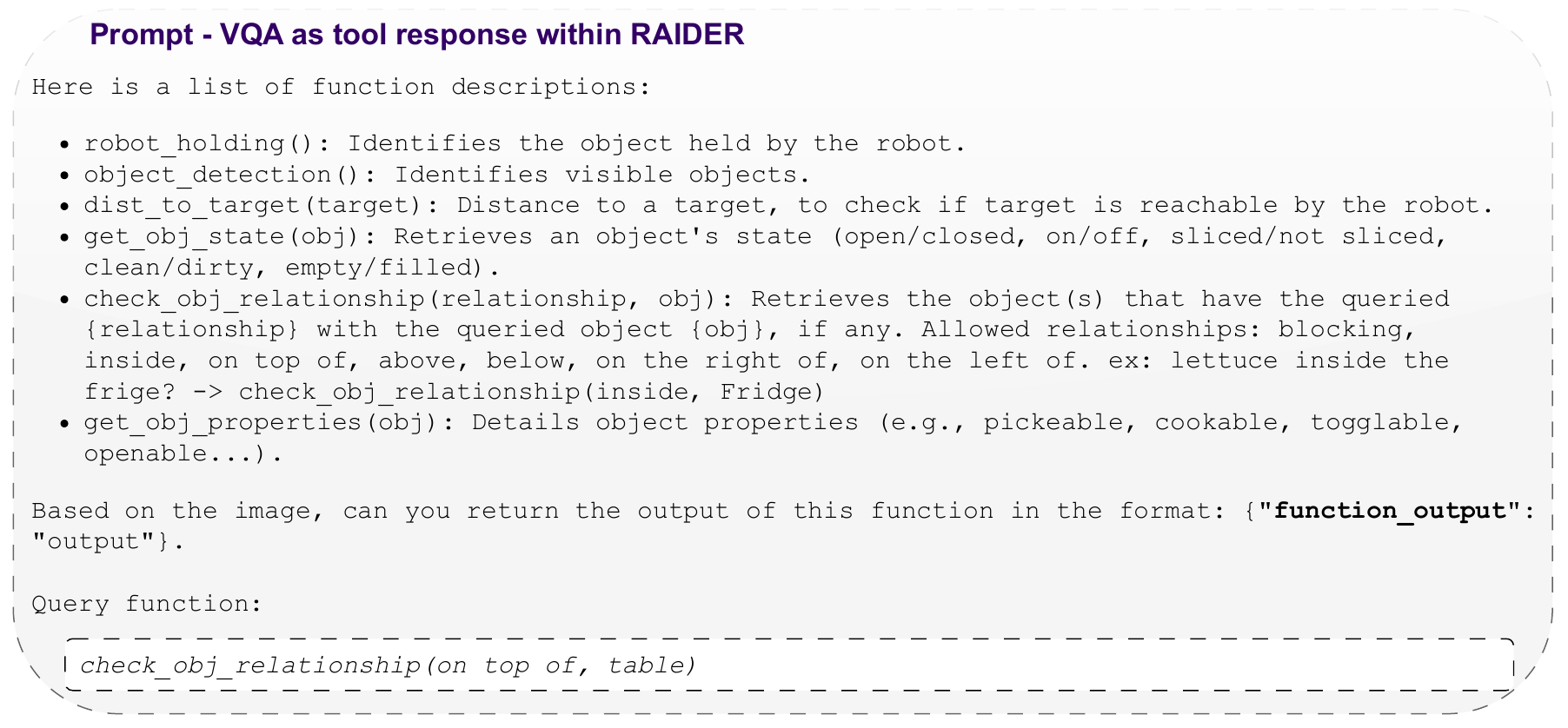}
    \caption{Prompt provided to the VQA model for tool call return.}
    \label{fig:prompt_llmvqa}
\end{figure}

In this method, the LLM agent tool implementations are replaced by the VQA model LLaVA-NeXT~\cite{liu2024llavanext}. The prompt provided to the VQA model when a tool is called by the agent is depicted in Figure~\ref{fig:prompt_llmvqa}. The rest of the system is not modified, including the prompt provided to the LLM, the LLM, and the program flow manager.

\section{Program Flow Manager Warnings}\label{app4_warnings}
\begin{figure}[H]
    \centering
    \includegraphics[width=\columnwidth]{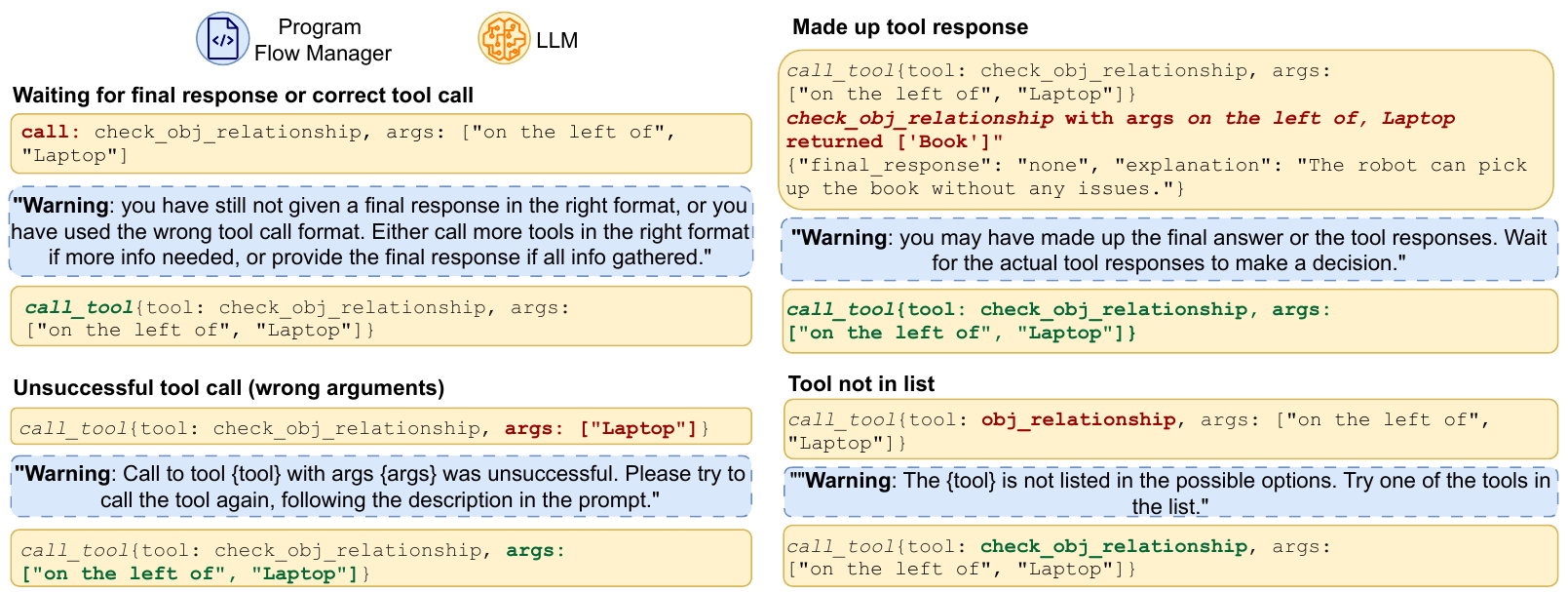}
    \caption{The Program Flow Manager detects incorrect tool calls or LLM responses, and issues warning messages to the LLM, prompting it to self-correct.}
    \label{fig:warnings}
\end{figure}

\section{RAIDER's outputs from Real-world Assistive Plan}\label{app5_demo}

We provide the full output generated by RAIDER for each action in the execution of the assistive plan described in Section~\ref{sec:extend_assist}, containing the complete flow between the LLM and the tools for issue detection and explanation.

\definecolor{usercolor}{RGB}{0, 102, 204} 
\definecolor{toolcolor}{RGB}{204, 0, 0}   
\definecolor{agentcolor}{RGB}{0, 153, 0}  

\begin{tcolorbox}[colframe=black, colback=white, title=1 - approach medicine\_counter. Unfeasibility: blocked path., breakable]

{\color{usercolor} \textbf{USER:}} approach medicine\_counter

{\color{agentcolor} \textbf{LLM:}} Step 1: Instance Grounding:  
- The action is to ``approach" the ``medicine\_counter."  
- I need to identify the ``medicine\_counter" in the environment.  

To proceed, I will use the object detection tool to identify the objects present in the scene and determine if there is a specific instance of a ``medicine\_counter."

Step 2: Question Generation:  
- Once the ``medicine\_counter" is identified, I need to check if the path to it is free and can be approached.  

I will generate the necessary tool calls to gather this information.

Step 3: Tool Calls - Information Gathering:  
- Call the object detection tool to identify the objects in the scene.  
- If the ``medicine\_counter" is identified, check if the path to it is free.  

Let's start with the object detection tool call.

\texttt{call\_tool\{"tool": "object\_detection", "args": []\}}

{\color{toolcolor} \textbf{TOOL CALL RESPONSE:}} \{'role': `user', `content': "Call to tool object\_detection with args [] returned ['medicine\_counter']"\}

{\color{agentcolor} \textbf{LLM:}} Step 1: Instance Grounding:  
- The ``medicine\_counter" has been identified in the environment.  

Step 2: Question Generation:  
- Now, I need to check if the path to the ``medicine\_counter" is free and can be approached.  

Step 3: Tool Calls - Information Gathering:  
- I will call the tool to check if the path to the ``medicine\_counter" is free.  

\texttt{call\_tool\{"tool": "check\_free\_path", "args": ["medicine\_counter"]\}}

{\color{toolcolor} \textbf{TOOL CALL RESPONSE:}} \{'role': `user', `content': ``Call to tool check\_free\_path with args ['medicine\_counter'] returned False"\}

{\color{agentcolor} \textbf{LLM:}} Step 5: Decision Making:  
- The path to the ``medicine\_counter" is not free, which means the robot cannot approach it without clearing the path first.  

Based on this information, the action is currently unfeasible.  

\texttt{\{"final\_response": "unfeasibility", "explanation": "The path to the medicine\_counter is not free, making it unapproachable without clearing the path."\}}

\end{tcolorbox}

\begin{tcolorbox}[colframe=black, colback=white, title=2 - pick adrianas\_medicine medicine\_counter. Ambiguity: two medicines., breakable]

{\color{usercolor} \textbf{USER:}} pick adrianas\_medicine medicine\_counter

{\color{agentcolor} \textbf{LLM:}} Step 1: Instance Grounding  
- Objects: The query involves picking an object, specifically ``adrianas\_medicine" from the ``medicine\_counter".  
- To ground these objects, I will use the object detection tool to identify the objects present in the scene and check their locations.  

Let's start by detecting the objects in the scene to identify ``adrianas\_medicine" and ``medicine\_counter".  

\texttt{call\_tool\{"tool": "object\_detection", "args": []\}}

{\color{toolcolor} \textbf{TOOL CALL RESPONSE:}} \{'role': `user', `content': ``Call to tool object\_detection with args [] returned [`medicine1', `medicine2', `plant', `bottle', `medicine\_counter']"\}

{\color{agentcolor} \textbf{LLM:}} Based on the object detection results, we have the following objects: `medicine1', `medicine2', `plant', and `bottle'.  

Step 1: Instance Grounding (continued)  
- Objects: ''adrianas\_medicine" is likely referring to either `medicine1' or `medicine2'. However, without additional context, there is ambiguity in determining which specific medicine is intended.  

\texttt{\{"final\_response": "ambiguity", "explanation": "The term `medicine' could refer to either `medicine1' or `medicine2', leading to ambiguity in identifying the specific object to pick."\}}

\end{tcolorbox}
\begin{tcolorbox}[colframe=black, colback=white, title=2 - pick adrianas\_medicine medicine\_counter. User info: ``It's the medicine that is closest to the plant.". Unfeasibility: not within reach., breakable]

{\color{usercolor} \textbf{USER:}} pick adrianas\_medicine medicine\_counter. ``It's the medicine that is closest to the plant.''

{\color{agentcolor} \textbf{LLM:}} Step 1: Instance Grounding  
- Objects: The action involves picking up ``adrianas\_medicine" from the ``medicine\_counter". The specific medicine is described as the one closest to the plant. I need to identify the plant and the medicines on the counter to determine which medicine is closest to the plant.  
- People: No direct interaction with a human is specified, but the action is for ``adrianas\_medicine", which might imply a human named Adriana is involved. However, since the action is picking an object, the focus is on object identification.  

Step 2: Question Generation  
- Is there a plant on the medicine counter, and what are the positions of the medicines relative to it?  
- Is the robot currently holding any object?  
- Is the medicine within 0.5m of the robot?  

Step 3: Tool Calls - Information Gathering  
I will use the following tools to gather the necessary information:  
\texttt{call\_tool\{"tool": "object\_detection", "args": []\}} to identify the plant and medicines on the counter.  
\texttt{call\_tool\{"tool": "robot\_holding", "args": []\}} to check if the robot is currently holding an object.  
\texttt{call\_tool\{"tool": "dist\_robot\_to\_obj", "args": ["adrianas\_medicine"]\}} to check the distance to the medicine.  

{\color{toolcolor} \textbf{TOOL CALL RESPONSE:}}  
\{'role': `user', `content': ``Call to tool object\_detection with args [] returned ['medicine1', `medicine2', `plant', `bottle']"\}  

{\color{toolcolor} \textbf{TOOL CALL RESPONSE:}}  
\{'role': `user', `content': `Call to tool robot\_holding with args [] returned None'\}  

{\color{toolcolor} \textbf{TOOL CALL RESPONSE:}}  
\{'role': `user', `content': ``Call to tool dist\_robot\_to\_obj with args ['adrianas\_medicine'] returned 0.6"\}  

{\color{agentcolor} \textbf{LLM:}} Step 1: Instance Grounding (continued)  
- Objects: There are two medicines detected: ``medicine1" and ``medicine2". The plant is also detected, but I need to determine which medicine is closest to the plant to identify ``adrianas\_medicine".  

Step 2: Question Generation (continued)  
- What is the distance between the plant and each of the medicines to determine which one is closest?  

Step 3: Tool Calls - Information Gathering (continued)  
I will use the following tool calls to gather the necessary information:  
\texttt{call\_tool\{"tool": "dist\_between\_objs", "args": ["plant", "medicine1"]\}}  
\texttt{call\_tool\{"tool": "dist\_between\_objs", "args": ["plant", "medicine2"]\}}  

{\color{toolcolor} \textbf{TOOL CALL RESPONSE:}}  
\{'role': `user', `content': ``Call to tool dist\_between\_objs with args ['plant', `medicine1'] returned 0.1"\}  

{\color{toolcolor} \textbf{TOOL CALL RESPONSE:}}  
\{'role': `user', `content': ``Call to tool dist\_between\_objs with args ['plant', `medicine2'] returned 0.2"\}  

{\color{agentcolor} \textbf{LLM:}} Step 1: Instance Grounding (conclusion)  
- Objects: ``medicine1" is the closest to the plant with a distance of 0.1, so it is identified as ``adrianas\_medicine".  

Step 2: Question Generation (finalization)  
- The robot is not currently holding any object, so it can pick up ``medicine1".  
- The distance to ``medicine1" is 0.6m, which is beyond the robot's picking range of 0.5m.  

Step 5: Decision Making  
- The action is unfeasible because the medicine is beyond the robot's picking range.  

Final response:  
\texttt{\{"final\_response": "unfeasibility", "explanation": "The medicine is 0.6m away, beyond the robot's picking range of 0.5m."\}}  

\end{tcolorbox}
\begin{tcolorbox}[colframe=black, colback=white, title=3 - approach adriana\_user. Unfeasibility: user not recognized., breakable]

{\color{usercolor} \textbf{USER:}} approach adriana\_user

{\color{agentcolor} \textbf{LLM:}} Step 1: Instance Grounding:  
- The action is to ``approach" a human named ``adriana\_user."  
- I need to identify if ``adriana\_user" is present in the environment.  

Step 2: Question Generation:  
- Is ``adriana\_user" present in the environment?  
- Is the path to ``adriana\_user" free and can be approached?  

Step 3: Tool Calls - Information Gathering:  
I will use the following tools to gather the necessary information:  
- \texttt{recognize\_humans()} to identify if "adriana\_user" is present.  
- \texttt{check\_free\_path(target)} to ensure the path to "adriana\_user" is clear.  

Let's start by identifying if ``adriana\_user" is present.  

\texttt{call\_tool\{"tool": "recognize\_humans", "args": []\}}  

{\color{toolcolor} \textbf{TOOL CALL RESPONSE:}}  
\{'role': `user', `content': ``Call to tool recognize\_humans with args [] returned ['-1']"\}  

{\color{agentcolor} \textbf{LLM:}} Step 1: Instance Grounding:  
- The tool response indicates that no specific humans, including ``adriana\_user," are recognized in the environment.  

Step 5: Decision Making:  
- Since ``adriana\_user" cannot be identified or recognized in the environment, the action to approach her is not feasible.  

Final response:  
\texttt{\{"final\_response": "unfeasibility", "explanation": "Adriana\_user is not recognized in the environment, making it impossible to approach her."\}}  

\end{tcolorbox}

\begin{tcolorbox}[colframe=black, colback=white, title=handover adrianas\_medicine adriana\_user. Unfeasibility: user is not looking., breakable]

{\color{usercolor} \textbf{USER:}} handover adrianas\_medicine adriana\_user

{\color{agentcolor} \textbf{AGENT:}} Step 1: Instance Grounding  
- Objects: The action involves handing over "adrianas\_medicine." I need to check if the robot is holding this specific object.  
- People: The action involves handing over the object to "adriana\_user." I need to identify if Adriana is present in the surroundings.  

Let's start by checking what the robot is currently holding and identifying Adriana in the surroundings.  

\texttt{call\_tool\{"tool": "robot\_holding", "args": []\}}

\texttt{call\_tool\{"tool": "recognize\_humans", "args": []\}}  

{\color{toolcolor} \textbf{TOOL CALL RESPONSE:}}  
\{'role': 'user', 'content': 'Call to tool robot\_holding with args [] returned medicine'\}  

{\color{toolcolor} \textbf{TOOL CALL RESPONSE:}}  
\{'role': 'user', 'content': "Call to tool recognize\_humans with args [] returned ['Adriana']"\}  

{\color{agentcolor} \textbf{AGENT:}} Step 1: Instance Grounding  
- Objects: The robot is holding "medicine," which matches "adrianas\_medicine."  
- People: "Adriana" is present in the surroundings, matching "adriana\_user."  

Step 2: Question Generation  
- Action: The action is to hand over the object.  
- Questions:  
  1. Is Adriana looking at the robot, indicating readiness for interaction?  
  2. Are Adriana's hands free to receive the object?  
  3. Is the distance between the robot and Adriana appropriate for a handover (less than 0.5m)?  

Let's gather this information using the appropriate tools.  

\texttt{call\_tool\{"tool": "detect\_human\_gaze", "args": ["Adriana"]\}}

\texttt{call\_tool\{"tool": "human\_hands\_free", "args": ["Adriana"]\}}

\texttt{call\_tool\{"tool": "dist\_robot\_to\_human", "args": ["Adriana"]\}}  

{\color{toolcolor} \textbf{TOOL CALL RESPONSE:}}  
\{'role': 'user', 'content': "Call to tool detect\_human\_gaze with args ['Adriana'] returned False"\}  

{\color{toolcolor} \textbf{TOOL CALL RESPONSE:}}  
\{'role': 'user', 'content': "Call to tool human\_hands\_free with args ['Adriana'] returned True"\}  

{\color{toolcolor} \textbf{TOOL CALL RESPONSE:}}  
\{'role': 'user', 'content': "Call to tool dist\_robot\_to\_human with args ['Adriana'] returned 0.4m"\}  

{\color{agentcolor} \textbf{AGENT:}} Step 4: Iterate Step 2 and Step 3  

Based on the gathered information:  
1. Adriana is not looking at the robot, indicating she may not be ready for interaction.  
2. Adriana's hands are free.  
3. The distance between the robot and Adriana is appropriate for a handover.  

Step 5: Decision Making  

Given that Adriana is not looking at the robot, it suggests she is not ready for the handover interaction.  

Final response:  
\texttt{\{"final\_response": "unfeasibility", "explanation": "Adriana is not looking at the robot, indicating she is not ready for interaction."\}}  

\end{tcolorbox}





\end{document}